\documentclass[10pt,twocolumn,letterpaper]{article}

\usepackage{cvpr}              
\usepackage{makecell} 
\usepackage{adjustbox}
\usepackage{colortbl}
\usepackage{multirow}
\usepackage{xcolor}
\definecolor{cvprblue}{rgb}{0.21,0.49,0.74}
\usepackage[pagebackref,breaklinks,colorlinks,allcolors=cvprblue,urlcolor=magenta]{hyperref}
\definecolor{highlight}{gray}{0.9}

\def\paperID{16010} 
\def\confName{CVPR}
\def\confYear{2026}

\title{V-ReasonBench: Toward Unified Reasoning Benchmark Suite for \\ Video Generation Models}

\author{\textbf{Yang Luo$^{1,*}$, Xuanlei Zhao$^{1,*}$, Baijiong Lin$^{2,*}$, Lingting Zhu$^{3}$, Liyao Tang$^{4}$\vspace{0.07cm}} \\ 
\textbf{Yuqi Liu$^{5}$, Ying-Cong Chen$^{2}$, Shengju Qian$^{6,\dagger}$, Xin Wang$^{6}$, Yang You$^{1}$\vspace{0.35cm}}\\
$^1$NUS \quad $^2$HKUST(GZ) \quad $^3$HKU \quad $^4$USYD  \quad  $^5$CUHK \quad $^6$LIGHTSPEED  \\ 
\mbox{}\\[0.05cm]
\centerline{Project Page: \url{https://oahzxl.github.io/VReasonBench/}}
}

\begin{document}
\maketitle
\def\thefootnote{*}\footnotetext{Equal Contribution.} \def\thefootnote{$\dagger$}\footnotetext{Corresponding Author.}
\begin{abstract}
Recent progress in generative video models, such as Veo-3, has shown surprising zero-shot reasoning abilities, creating a growing need for systematic and reliable evaluation. We introduce V-ReasonBench, a benchmark designed to assess video reasoning across four key dimensions: structured problem-solving, spatial cognition, pattern-based inference, and physical dynamics. The benchmark is built from both synthetic and real-world image sequences and provides a diverse set of answer-verifiable tasks that are reproducible, scalable, and unambiguous. Evaluations of six state-of-the-art video models reveal clear dimension-wise differences, with strong variation in structured, spatial, pattern-based, and physical reasoning. We further compare video models with strong image models, analyze common hallucination behaviors, and study how video duration affects Chain-of-Frames reasoning. Overall, V-ReasonBench offers a unified and reproducible framework for measuring video reasoning and aims to support the development of models with more reliable, human-aligned reasoning skills.
\end{abstract}    
\section{Introduction}
\label{sec:intro}

Recent advancements in generative video models \citep{HaCohen2024LTXVideo, zhang2025show1marryingpixellatent,polyak2025moviegencastmedia, chen2025skyreelsv2infinitelengthfilmgenerative, ma2025stepvideot2vtechnicalreportpractice}, such as Veo-3.1~\citep{google2025veo3}, Sora-2 \citep{openai2025sora2}, and Kling-2.5 \citep{kuaishou2024kling}, have demonstrated remarkable zero-shot reasoning capabilities across diverse visual tasks. These models exhibit emergent abilities to perceive, model, and reason over dynamic visual data, reflecting a growing trajectory toward general-purpose vision foundation models. Such progress suggests that video models are beginning to move beyond mere frame synthesis, displaying cognitive-like behaviors such as pattern inference, causal prediction, and strategic planning \citep{Wiedemer2025VideoZeroShot}. However, despite these promising signs, the capacity to quantitatively evaluate reasoning ability in these models remains underexplored and poorly standardized.

\begin{figure}[!t]
    \centering
    \includegraphics[width=.9\linewidth]{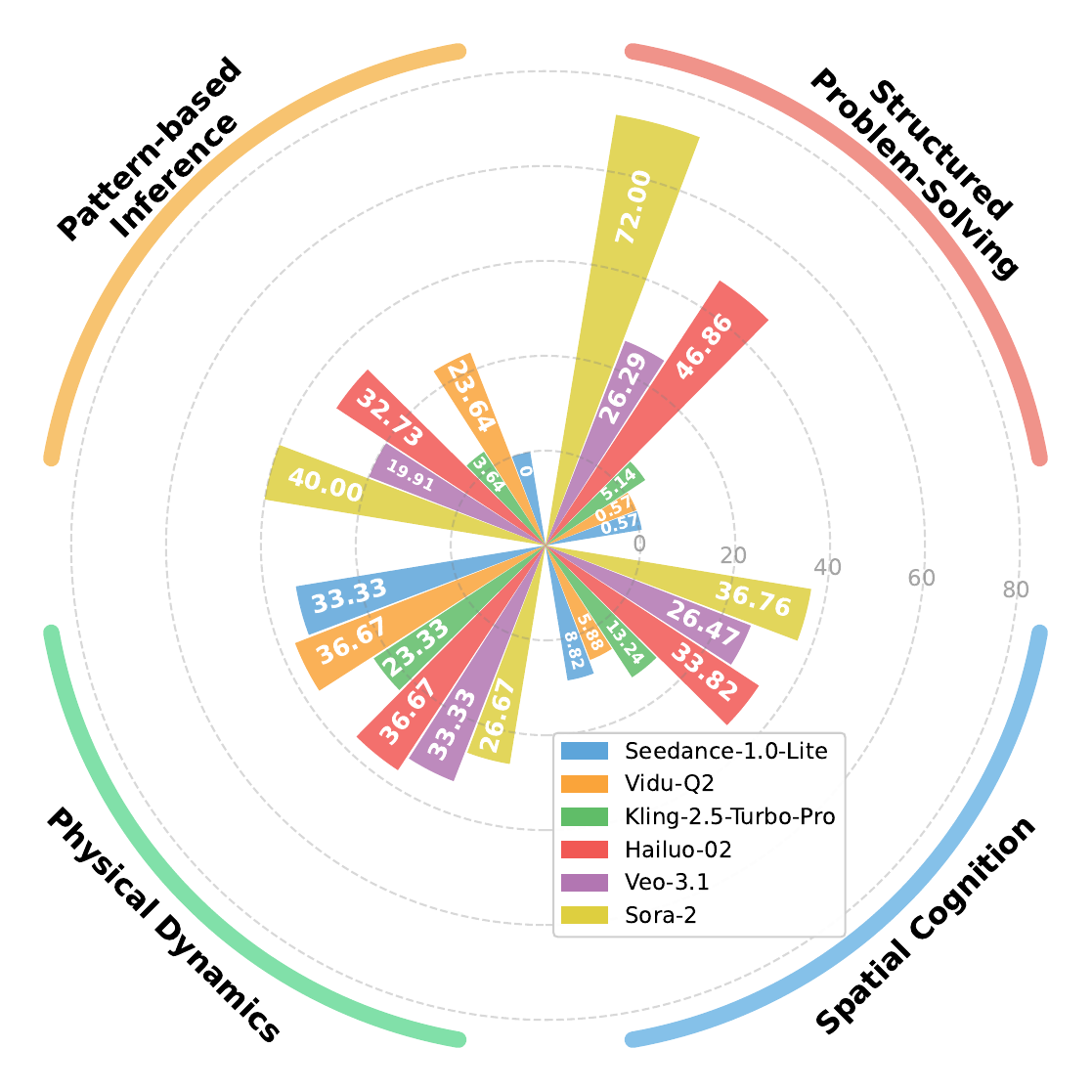}
    \caption{Evaluation of Video Generation Models on V-ReasonBench.
The performance of six video generation models across the four core reasoning dimensions is illustrated. Detailed numerical results are provided in Tab. \ref{tab:model_dimension}.}
    \label{fig:visual_results}
      \vspace{-10pt}
\end{figure}

The “Chain-of-Frame” (CoF) \citep{Wiedemer2025VideoZeroShot, Ghazanfari2025CoF} paradigm treats video generation as a sequence of reasoning steps, in direct analogy to “Chain-of-Thought” in language models \citep{Wei2022CoT, Zhou2022LeastToMost, Wang2022SelfConsistency, wang2025videorftincentivizingvideoreasoning, li2025videochatr1enhancingspatiotemporalperception}. In this framework, a model receives an initial image and a prompt, then produces a series of frames where the intermediate frames embody its reasoning trajectory and the final frame represents its answer or outcome. Because the final frame encapsulates the model’s inferred solution, CoF enables a last-frame evaluation pipeline: we judge the model on its concluding frame rather than requiring annotation of all intermediate steps. This provides an efficient and scalable approach for unambiguous evaluation of video reasoning, because evaluating every intermediate step requires substantial annotation effort, incurs prohibitive computational cost, and often yields noisy or ambiguous supervisory signals \citep{chu2024navigate, kamoi2025visonlyqalargevisionlanguage, shinde2025surveyefficientvisionlanguagemodels}.

While last-frame evaluation offers an efficient way to assess reasoning within the CoF paradigm, vision-language models (VLMs) are not always reliable as sole automatic judges. In particular, VLMs may face difficulty interpreting visually dense or grid-structured layouts that require precise recognition of small cells, thin boundaries, or subtle geometric relationships \cite{Li2024TopViewSpatial,Pothiraj2025CAPTURE, kamoi2025visonlyqalargevisionlanguage}. To complement these limitations and ensure stable evaluation across diverse task types, V-ReasonBench adopts a hybrid strategy: mask-based evaluation for tasks with well-defined object regions, grid-based evaluation for tasks requiring fine-grained structural accuracy, and lightweight VLM-based evaluation only for visually simple outputs where consistency can be maintained. Each method produces a numerical score that is converted into a pass or fail decision using task-specific thresholds, enabling scalable and reproducible pass@k evaluation while preserving strong alignment with human judgment.

To support consistent evaluation under the Chain-of-Frame framework and our last-frame scoring strategy, V-ReasonBench is partitioned into four complementary reasoning classes, each targeting a specific dimension of reasoning: 1) Structured problem-solving encompasses tasks involving numerical manipulation, strategic planning in dynamic game states, and procedural logic derived from visualized program traces. 2) Spatial cognition evaluates understanding of spatial relations, geometric transformations, and symmetry-based patterns. 3) Pattern-based inference probes sequence completion, analogical mapping, and abstract rule induction beyond surface-level visual cues. 4) Physical dynamics examines comprehension of motion, force interactions, temperature effects, and pressure-driven behavior. Together, these four classes provide a comprehensive assessment of cognitive capabilities in video understanding and reasoning.

Built upon these methodological principles, V-ReasonBench provides a unified suite for evaluating reasoning in generative video models under the CoF framework. Each instance consists of an initial image, a task instruction, and a target final image, with both procedurally generated and curated scenarios enabling fine control over difficulty and perceptual variation. By applying deterministic, last-frame scoring across four complementary reasoning classes, the benchmark offers consistent measurements of a model’s ability to interpret structured inputs, follow task-specific rules, and produce coherent final outcomes. 

Evaluating six cutting-edge video generation models (Sora-2 \citep{openai2025sora2}, Veo-3.1 \citep{google2025veo3}, Kling-2.5-Turbo-Pro \citep{kuaishou2024kling}, Seedance-1.0-Lite \citep{gao2025seedance}, Vidu-Q2 \citep{shengshu2025viduq2}, and Hailuo-02 \citep{minimax2025hailuo02}) reveals clear strengths and limitations across multiple reasoning dimensions in Fig. \ref{fig:visual_results}. Our analysis further examines each model’s characteristic reasoning patterns, the influence of video duration on CoF performance, and the contrast between video and image models, highlighting several recurring hallucination behaviors unique to video generation. As an early exploratory effort, V-ReasonBench provides an efficient and reproducible reasoning-centric framework that offers a unified foundation for benchmarking video reasoning and guiding future models toward more reliable, human-aligned visual reasoning.
\section{Related Works}
\label{sec:related}

\noindent\textbf{Video Generation.}
Recent advances in video generation have been strongly driven by diffusion–transformer models, which provide scalable architectures for producing high-quality visual content~\citep{chen2024gentrondiffusiontransformersimage, ma2024lattelatentdiffusiontransformer, gao2024luminat2xtransformingtextmodality, zheng2024opensorademocratizingefficientvideo}. Commercial systems such as OpenAI’s Sora-2~\citep{openai2025sora2}, Minimax’s Hailuo-02~\citep{minimax2025hailuo02}, Runway’s Gen-3~\citep{runway2024gen3}, and Google DeepMind’s Veo-3.1~\citep{google2025veo3} achieve impressive visual quality and strong alignment with text prompts, but their training data, model designs, and evaluation procedures remain undisclosed. Alongside these developments, recent research efforts, including CogVideoX~\citep{yang2025cogvideoxtexttovideodiffusionmodels}, HunyuanVideo~\citep{tencent2024hunyuanvideo}, and the Wan series~\citep{wan2025wan}, have focused on enhancing temporal consistency, motion fidelity, and semantic coherence, reflecting a broader trend toward more controllable and robust video generation systems.

\noindent\textbf{Video Reasoning.}
Recent video generation models have shown emergent visual reasoning abilities that go beyond their original training goals~\citep{Wiedemer2025VideoZeroShot}. The \emph{Chain-of-Frame} (CoF) paradigm~\citep{Wiedemer2025VideoZeroShot}, analogous to chain-of-thought reasoning in language models~\citep{wei2022chain}, frames video generation as a stepwise spatial and temporal reasoning process. This idea has been explored through approaches such as VChain~\citep{huang2025vchain}, which uses multimodal models to produce keyframes as intermediate reasoning signals, and visual chain-of-thought methods~\citep{chen2024visual,zhao2025cotvla}, which combine visual perception with iterative reasoning for knowledge-intensive tasks and control. However, despite growing interest in evaluating these emergent capabilities, there remains a lack of a scalable and comprehensive benchmark that systematically measures the breadth and depth of video reasoning across diverse task settings.

\begin{figure*}[ht]
    \centering
    \includegraphics[width=\linewidth]{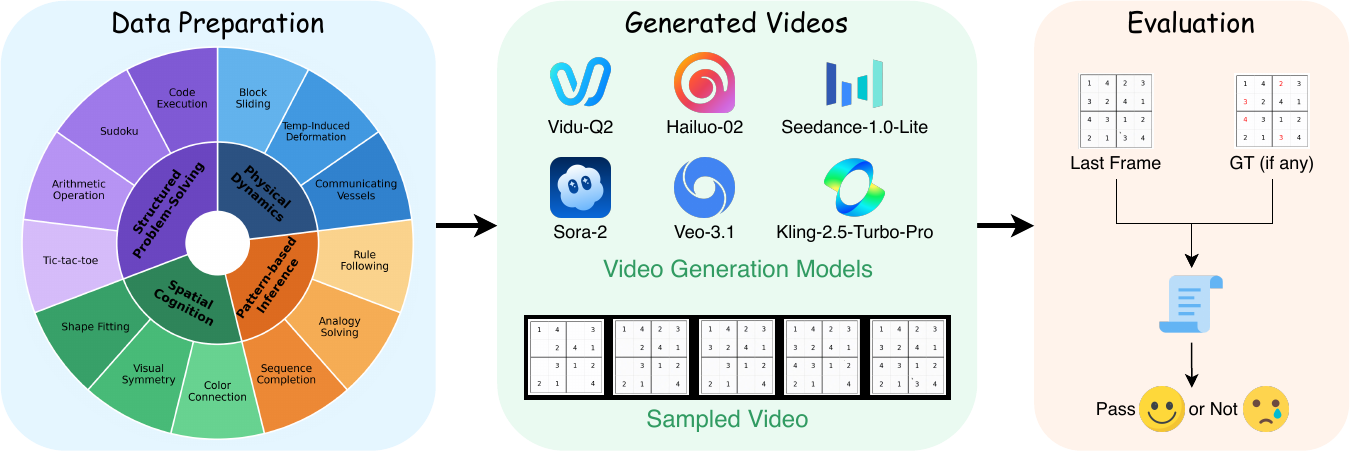}
    \caption{Overview of V-ReasonBench pipeline. The benchmark covers four reasoning dimensions, integrates both synthetic and real-world scenarios, and supports reproducible, large-scale evaluation of video reasoning capabilities.}
    \label{fig:pipeline}
\end{figure*}

\section{V-ReasonBench Pipeline}
\label{sec:vr}

\subsection{Design Principles}

V-ReasonBench is designed to evaluate the reasoning capabilities of generative video models through a comprehensive suite of tasks that require genuine understanding of visual reasoning, as illustrated in Fig. \ref{fig:pipeline}. Our benchmark is built on three core principles:

\begin{itemize}
    \item \textbf{Last-Frame Dependency:} All tasks are designed such that the final answer can be determined exclusively from the last frame of generated videos, enabling unambiguous and scalable evaluation of reasoning outcomes.
    \item \textbf{Unified Evaluation Metric:} We employ pass@k as our primary evaluation metric across all reasoning classes, providing a consistent measure of model performance and facilitating direct comparison between different approaches.
    \item \textbf{Multi-Faceted Reasoning:} The benchmark spans 4 distinct reasoning dimensions to comprehensively assess models' capabilities across mathematical, strategic, spatial, logical, physical, and programmatic reasoning.
\end{itemize}

\subsection{Reasoning Tasks and Formulations}

V-ReasonBench formalizes a comprehensive suite of reasoning classes, each targeting specific cognitive capabilities essential for holistic video understanding. The following is an overview of the 4 core reasoning dimensions in our benchmark. More explanations and visualizations of all tasks are provided in Appendix~\ref{appendix:details}.

\noindent\textbf{Structured Problem-Solving.}
This dimension evaluates the model’s ability to perform systematic, rule-based reasoning across quantitative and procedural contexts. It consists of four sub-tasks: 
\begin{itemize}
    \item \textbf{Arithmetic Operation} examines numerical reasoning from visual representations, requiring the model to perform basic arithmetic operations, including addition, subtraction, multiplication, and division.
    \item \textbf{Code Execution} tests procedural and logical thinking through visualized programming tasks, where the model observes LeetCode-style code with inputs and must predict the correct program output.
    \item \textbf{Sudoku} evaluates constraint-based reasoning by asking the model to complete a partially filled grid while adhering to fixed numeric rules.
    \item \textbf{Tic-Tac-Toe} assesses strategic planning in a dynamic, adversarial environment, requiring the model to infer or select the optimal next move given a game state.
\end{itemize}

\begin{figure*}
    \centering
    \includegraphics[width=\linewidth]{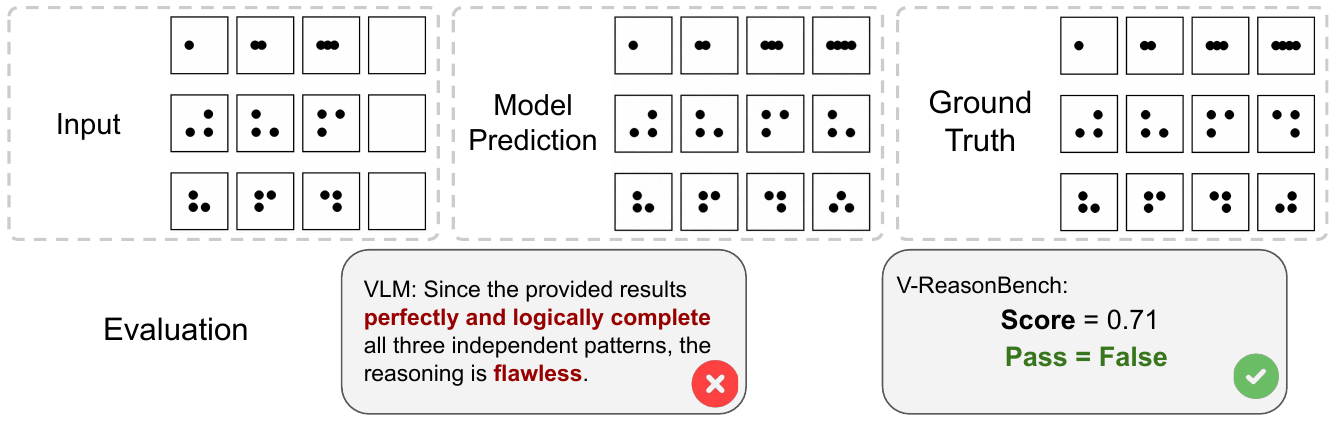}
    \caption{Example failure case from \textit{Sequence Completion} task illustrating the limitations of VLM-based automatic evaluation. Although the underlying rule is simple, the VLM incorrectly assesses the model’s output due to difficulties in recognizing small grid cells and fine structural differences. More examples are given in Appendix~\ref{appendix:vlm}.}
    \label{fig:vlm}
\end{figure*}
\noindent\textbf{Spatial Cognition.}
This dimension examines spatial intelligence across three sub-tasks: 
\begin{itemize}
    \item \textbf{Shape Fitting} assesses mental rotation and spatial arrangement skills.
    \item \textbf{Visual Symmetry} evaluates recognition of reflective and rotational symmetries.
    \item \textbf{Color Connection} tests pathfinding by requiring the model to establish valid connections between elements of matching color.
\end{itemize}

\noindent\textbf{Pattern-based Inference.}
This dimension probes abstract relational reasoning and inductive generalization across three sub-domains:
\begin{itemize}
    \item \textbf{Sequence Completion} evaluates the ability to infer the next element in a visual or symbolic progression by recognizing hidden temporal or spatial rules.
    \item \textbf{Analogy Solving} tests the understanding of relational structure through problems of the form “A:B as C:?” requiring cross-domain correspondence beyond surface similarity.
    \item \textbf{Rule Following} examines the capacity to infer governing principles from a few examples and apply them consistently to novel instances.
\end{itemize}

\noindent\textbf{Physical Dynamics.}
This dimension evaluates the understanding of fundamental physical principles through four sub-tasks:  
\begin{itemize}
    \item \textbf{Block Sliding} examines the ability to predict motion under gravity and friction, determining whether an object placed on a slope will remain stationary or slide down.
    \item \textbf{Communicating Vessels (CV)} evaluates understanding of fluid pressure and equilibrium, asking the model to infer how liquid levels adjust across connected containers when pressure or volume changes.
    \item \textbf{Temperature-Induced Deformation} assesses reasoning about material properties under thermal variation, such as predicting melting, shrinking, or deformation of ice blocks as temperature changes.
\end{itemize}

\subsection{Dataset Construction}

\noindent\textbf{Image Pair Generation.}
We instantiate the benchmark with an image–pair framework rather than full videos. Approximately 90\% of instances are programmatically synthesized in custom simulation environments, yielding an \textit{initial image} that represents the starting state and a \textit{final image} that serves as the ground truth outcome, which enables scalable data generation. Procedural generation provides broad coverage across reasoning types while preserving consistent state transitions, and every pair passes automated validation followed by targeted manual inspection to ensure that each transformation is unambiguous, solvable by reasoning, and free of annotation shortcuts. This design follows best practices in controlled datasets for spatiotemporal and causal reasoning \citep{Yi2020CLEVRER,Girdhar2019CATER,GrundeMcLaughlin2021AGQA}. 

\begin{table}[t]
\centering
\caption{Overview of reasoning dimensions, tasks, and number of videos in V-ReasonBench. Each instance is an initial–final image pair, with each model generating five videos for pass@5 evaluation.}
\label{tab:vrb_stats}
\begin{tabular}{p{2.5cm} p{3cm} c}
\toprule
\textbf{Dimension} & \textbf{Task} & \textbf{\# Videos} \\
\midrule

\multirow{4}{*}{\parbox{3.2cm}{Structured\\Problem-Solving}}
    & Arithmetic Operation & \multirow{4}{*}{5,250} \\
    & Code Execution & \\
    & Sudoku & \\
    & Tic-tac-toe & \\

\midrule

\multirow{3}{*}{\parbox{3.2cm}{Spatial\\Cognition}}
    & Shape Fitting & \multirow{3}{*}{2,040} \\
    & Visual Symmetry & \\
    & Color Connection & \\

\midrule

\multirow{3}{*}{\parbox{3.2cm}{Pattern-based\\Inference}}
    & Sequence Complete & \multirow{3}{*}{1,590} \\
    & Analogy Solving & \\
    & Rule Follow & \\

\midrule

\multirow{3}{*}{\parbox{3.2cm}{Physical\\Dynamics}}
    & Block Sliding & \multirow{3}{*}{900} \\
    & CV & \\
    & Temperature & \\

\bottomrule
\end{tabular}
  \vspace{-5pt}
\end{table}
\noindent\textbf{Dataset Statistics.} V-ReasonBench includes 326 reasoning instances represented by 652 images paired with corresponding question annotations across four reasoning classes as summarized in Tab.~\ref{tab:vrb_stats}. By generating 5 videos for each instance-model pair, this yields a total of 9{,}780 generated videos assessed in the benchmark. Each instance is specified by an initial–final image pair that encodes the state transition necessary for reasoning.

\subsection{Limitations of VLM-Based Evaluation}
Although recent benchmarks increasingly rely on vision-language models (VLMs) for automatic judgment, this approach faces notable limitations. VLMs often struggle to interpret complex visual layouts, particularly grid-based or densely structured scenes that require precise detection of small cells, thin boundaries, or fine spatial relationships, as shown in Fig.~\ref{fig:vlm}. These difficulties lead to incorrect judgments even when the task logic is straightforward, as the model may miscount, overlook subtle structural cues, or misidentify local patterns. These challenges highlight the need for more reliable evaluation methods for tasks that depend on fine-grained visual understanding.

\begin{table*}[ht]
\centering
\small
\caption{Model-level overall and per-dimension performance on V-ReasonBench. A pass@5 score for each model is calculated within each dimension and presented accordingly.}
\begin{adjustbox}{width=0.8\textwidth}
\begin{tabular}{lcccccc}
\toprule
\makecell{\textbf{Model}} &
\makecell{{\textbf{Structured}}\\{\textbf{Problem-Solving}}} &
\makecell{{\textbf{Spatial}}\\{\textbf{Cognition}}} &
\makecell{{\textbf{Pattern-based}}\\{\textbf{Inference}}} &
\makecell{{\textbf{Physical}}\\{\textbf{Dynamics}}} &
\makecell{\textbf{Average}} \\
\cmidrule(r){1-1} \cmidrule{2-5} \cmidrule(l){6-6}
Seedance-1.0-Lite~\cite{gao2025seedance} & 0.57 & 8.82 & 0.00 & 33.33 & 10.68  \\
Vidu-Q2~\cite{shengshu2025viduq2} & 0.57 & 5.88 & 23.64 & \cellcolor{highlight}\textbf{36.67} & 16.69 \\
Kling-2.5-Turbo-Pro ~\cite{kuaishou2024kling} & 5.14 & 13.24 & 3.64 & 23.33 & 11.34  \\
Veo-3.1~\cite{google2025veo3} & 26.29 & 26.47 & 10.91 & 33.33 &  24.25  \\
Hailuo-02~\cite{minimax2025hailuo02} & 46.86 & 33.82 & 32.73 & \cellcolor{highlight}\textbf{36.67} & 37.52 \\
Sora-2~\cite{openai2025sora2} & \cellcolor{highlight}\textbf{72.00} & \cellcolor{highlight}\textbf{36.76} & \cellcolor{highlight}\textbf{40.00} & 26.67 &  \cellcolor{highlight}\textbf{43.86}  \\
\bottomrule
\end{tabular}
\label{tab:model_dimension}
\end{adjustbox}
\end{table*}
\subsection{Evaluation Methodology}

To enable precise and scalable last-frame-based evaluation, we design three complementary assessment methods tailored to the characteristics of different task types in V-ReasonBench.

\begin{itemize}

    \item \textbf{Mask-based Evaluation.}
    Tasks with clear object boundaries and localized reasoning regions, including \textit{Sequence Completion}, \textit{Analogy Solving}, \textit{Block Sliding}, \textit{Communicating Vessel}, and \textit{Tic-tac-toe}, are evaluated using a mask-based comparison strategy. Masks are generated either from annotated templates or through automated segmentation tools such as SAM-2 \cite{ravi2024sam2segmentimages} to isolate target regions. Pixel-level mean squared error is then computed with higher weights inside the masked areas and lower weights in non-essential regions, reducing the influence of background changes or stylistic variations.

    \item \textbf{Grid-based Evaluation.}
    For tasks that require structured layouts or fine-grained spatial precision, including \textit{Visual Symmetry}, and \textit{Rule Following}, we employ a grid-based evaluation. Each frame is divided into uniform cells, and cell-wise accuracy is computed by comparing the predicted and ground truth states in corresponding grid locations. This method captures discrete positional errors and ensures sensitivity to geometric and structural constraints.

    \item \textbf{VLM-based Evaluation.}
    Tasks composed of simple items that VLMs can easily handle, including \textit{Arithmetic Operation}, \textit{Sudoku}, \textit{Code Execution}, \textit{Temperature-Induced Deformation}, \textit{Color Connection}, and \textit{Shape Fitting}, are scored using a lightweight VLM-based procedure. For mathematical and code tasks, the VLM extracts textual or symbolic outputs from designated regions. For perception-oriented tasks, the VLM assesses structural correctness and relational consistency based on carefully designed prompts. This approach provides flexible and standardized scoring where purely pixel-based metrics are insufficient.

\end{itemize}

Each of these evaluation strategies produces a numerical score for the final frame, which is then converted into a binary \textbf{passed} or \textbf{unpassed} decision using task-specific thresholds. This unified procedure ensures consistent, scalable, and interpretable pass@k evaluation across all reasoning categories, with strong alignment to human assessment.

\section{Experiments}
\label{sec:expr}

\subsection{Experimental Setup}
\noindent\textbf{Models Evaluated.}
We evaluate six prevalent generative video models that represent the current frontier of commercial video generation capabilities. The models include: Sora-2 \citep{openai2025sora2}, Veo-3.1 \cite{google2025veo3}, Hailuo-02 \citep{minimax2025hailuo02}, Vidu-Q2 \cite{shengshu2025viduq2}, KlingAI-2.5-Turbo-Pro \citep{kuaishou2024kling}, and Seedance-1.0-Lite \citep{gao2025seedance}. This diverse selection supports a comprehensive evaluation of reasoning behaviors across models with different architectures, training pipelines, and capability tiers.

\noindent\textbf{Evaluation Protocol.}
We evaluate each model using a standardized protocol that generates five videos per prompt. All models are run with their default parameter settings and produce videos of approximately five seconds in duration. Resolutions of 720p or 768p are used when supported, and every video is generated in a 16:9 aspect ratio to ensure consistency. Video generation prompts are carefully designed and kept identical across models to minimize prompt-induced bias. Full prompts, score calculation methods and corresponding thresholds for each task are provided in Appendix~\ref{appendix:details}. For all tasks, we use pass@5 as the primary evaluation metric. 
In VLM-based evaluation tasks, such as color connect and shape fit tasks, where VLM-based assessment is reliable, we employ Gemini-2.5-Pro \cite{comanici2025gemini25pushingfrontier} as the automatic evaluator.

\subsection{Per-dimension Evaluation}
Across the four reasoning dimensions, the results in Tab.~\ref{tab:model_dimension} suggest that current video models exhibit distinct areas of strength rather than uniformly high performance. Sora-2 \citep{openai2025sora2} continues to lead in \textbf{Structured Problem-Solving} (72.00), \textbf{Spatial Cognition} (36.76), and \textbf{Pattern-based Inference} (40.00), with Hailuo-02 \citep{minimax2025hailuo02} following as a strong performer in these categories. These trends indicate that some models have developed comparatively stronger capabilities in discrete decision-making, spatial understanding, and abstraction from visual patterns. In contrast, Seedance-1.0-Lite \citep{gao2025seedance}, Vidu-Q2 \citep{shengshu2025viduq2}, and Kling-2.5-Turbo-Pro \citep{kuaishou2024kling} score lower in these dimensions, which may reflect differing design priorities, such as a stronger focus on visual generation fidelity over rule-driven or symbolic reasoning.

A different pattern emerges in \textbf{Physical Dynamics}. Here, Hailuo-02 and Vidu-Q2 achieve the highest scores (both 36.67), while Sora-2 reaches a moderate 26.67 despite its strengths in the other dimensions. This divergence suggests that the inductive biases supporting structured and pattern-based reasoning may not directly translate into robust physical understanding, and that some models may instead focus on producing visually coherent physical motions without fully capturing underlying physical principles. Overall, the dimension-wise results highlight that video reasoning is multifaceted: different systems capture different aspects of the task, and there remains substantial opportunity to develop models that more uniformly integrate abstract, spatial, and physically grounded reasoning.

\subsection{Human Preference Alignment}
\begin{figure}
    \centering
    \includegraphics[width=\linewidth]{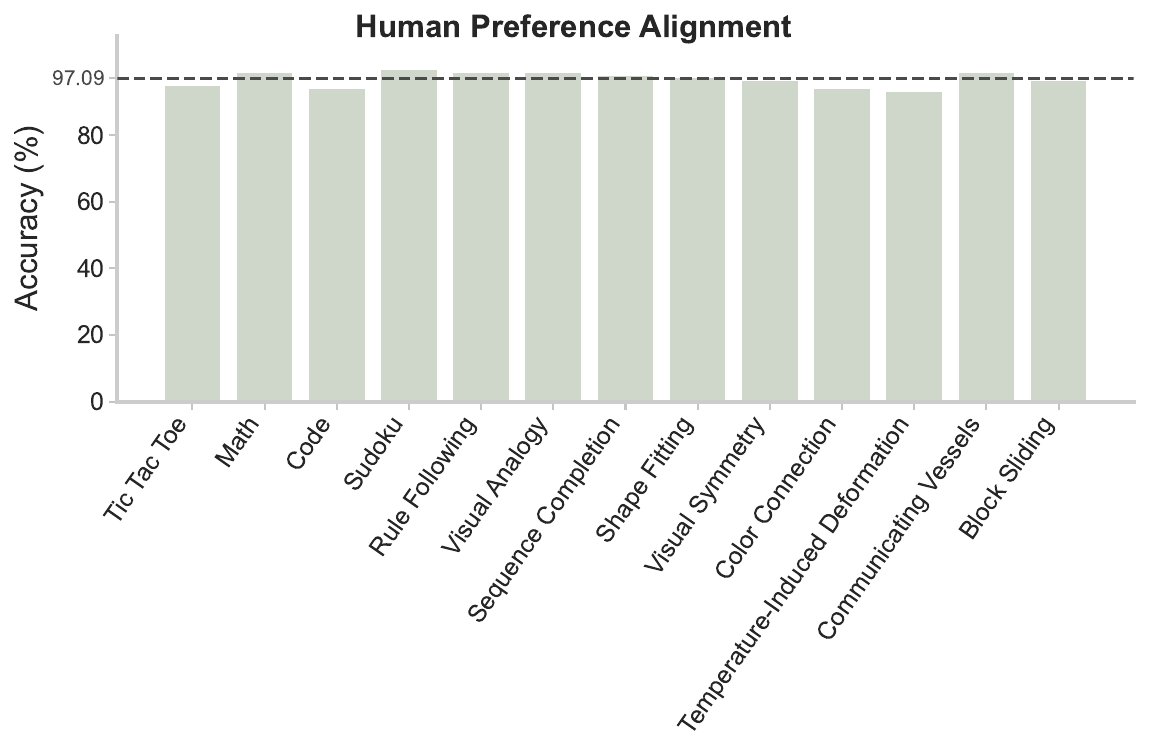}
    \caption{Human–alignment validation of our benchmark’s scoring pipeline. Each point compares binary \textbf{Pass/Unpass} decisions from the automatic evaluation with human judgments across four reasoning categories.}
    \label{fig:human}
\end{figure}
To verify the reliability of our scoring pipeline, we perform a human–alignment study that directly compares the binary passed/unpassed decisions produced by our benchmark with human judgments. For each reasoning task, we randomly sample 120 videos spanning a range of difficulty levels. Six graduate-level examiners familiar with reasoning benchmarks independently assess whether each model’s final-frame prediction correctly solves the task, following the same execution-based criteria defined in our benchmark.

We then compute the accuracy of agreement between the benchmark’s passed/unpassed outputs and the human passed/unpassed labels. Across models and categories, we observe consistently high alignment (accuracy = 97.09\% on average) as presented in Fig.~\ref{fig:human}, demonstrating that our automatic decisions closely track human evaluations. The remaining discrepancies typically arise in visually ambiguous cases, such as nearly symmetric configurations or physics scenarios with partial occlusion, where human evaluators show slightly higher tolerance to minor perceptual deviations.

Overall, these results indicate that our evaluation protocol produces human-consistent pass/fail judgments and that V-ReasonBench provides a scalable, reproducible assessment framework that remains well aligned with human reasoning preferences.

\section{Discussions}
\label{sec:discussions}


\subsection{Reasoning Patterns in Video Generation}
\begin{figure}
    \centering
    \includegraphics[width=\linewidth]{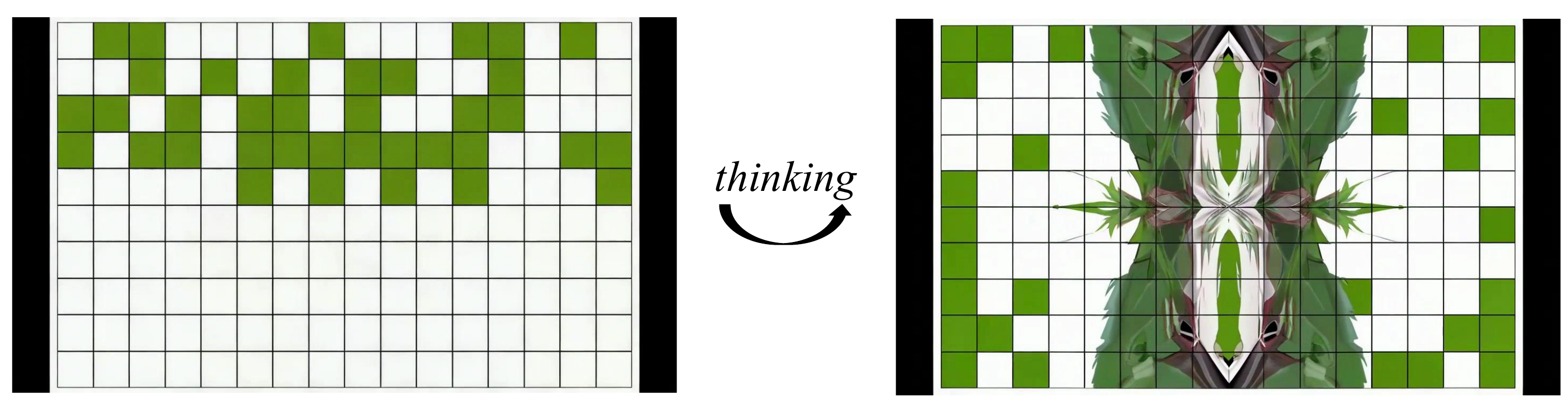}
    \caption{Example from the Seedance-1.0-Lite model on the horizontal visual symmetry task. The model introduces additional decorative patterns across the mirrored axis, illustrating its tendency to enrich visual appearance rather than preserve original geometric form.}
    \label{fig:demo_pattern}
\end{figure}
Across several reasoning dimensions, we observe a consistent phenomenon in which some video models emphasize visual enhancement over structural accuracy, which correlates with lower reasoning performance. Taking Seedance-1.0-Lite as examples in Fig. \ref{fig:demo_pattern}, the tendency appears on tasks featuring minimal, clean backgrounds (often pure white or black) with only the required elements.
In these minimalist settings, the models frequently add extra textures or objects or modify the scene layout, which alters the intended structure and reduces pass@5 scores. The same pattern appears beyond structured problem solving, including spatial and geometric tasks, pattern-based inference, and parts of physical reasoning that require a stable canvas. More visualizations of reason patterns can be found in Appendix~\ref{appendix:reason}.

A likely cause is a creative bias inherited from pretraining on large, open-domain video corpora where realism and visual richness are valued. When the input scene is visually sparse, the model may treat it as incomplete and attempt to improve it with additional detail rather than preserve the given structure. Training and decoding choices can strengthen this behavior, such as reconstruction objectives that reward fine texture, temporal smoothness terms that encourage motion even when the correct solution is static, and limited exposure to diagram-like data with small symbols and thin boundaries. Together, these factors push the generator toward aesthetic completion instead of structure-preserving rendering, which can conflict with tasks that rely on precise spatial or symbolic constraints.

\subsection{Influence of Duration on Video Reasoning}
\label{subsec:res_dur}

\begin{figure}[!t]
    \centering
    \includegraphics[width=\linewidth]{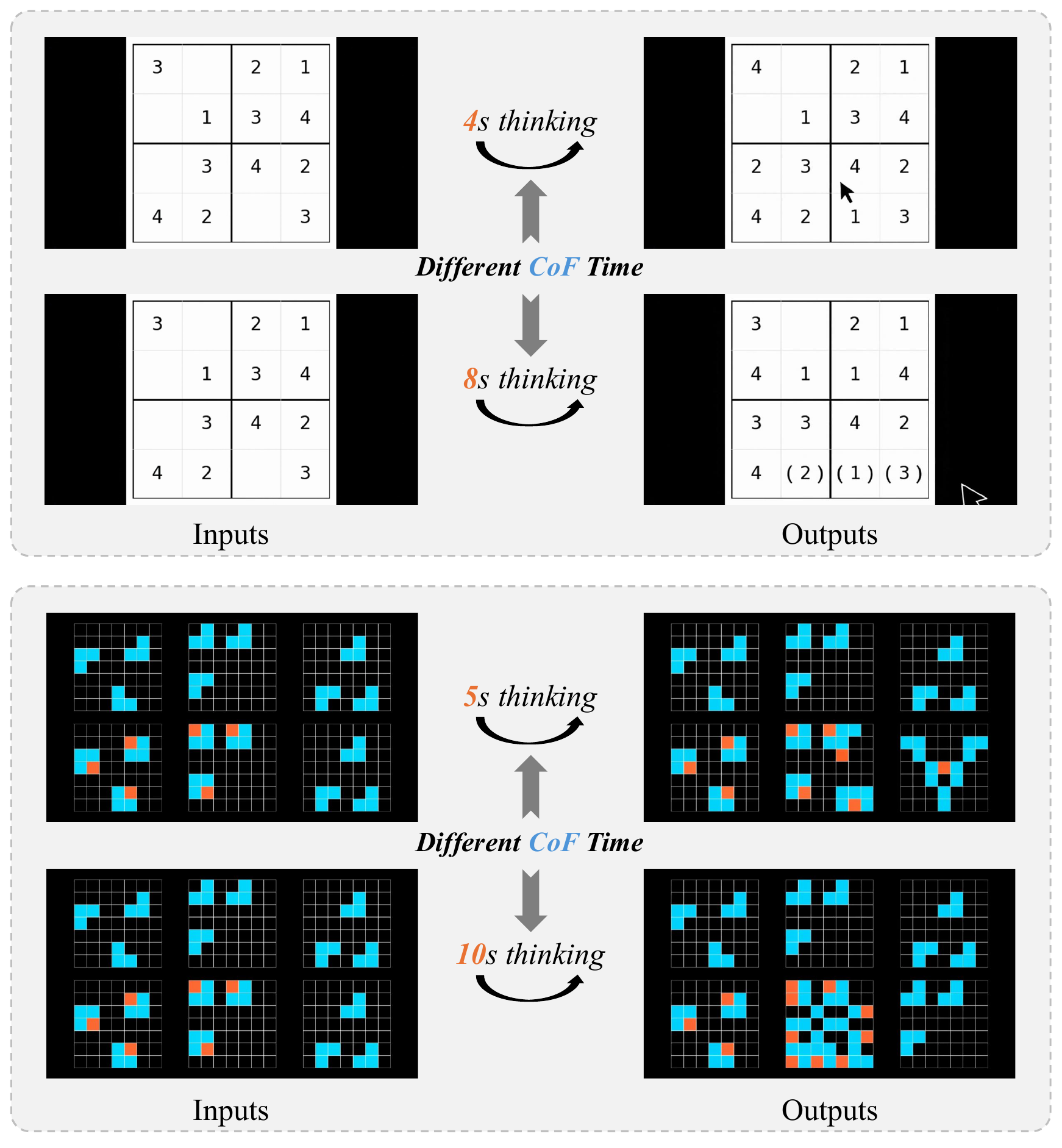}
    \caption{Effect of video duration on reasoning outcomes of Sora-2 in the Chain-of-Frame setting. Each row compares model generations with different “thinking” durations for tasks such as \textit{Sudoku} and \textit{Rule Following}. Although longer durations correspond to longer reasoning processes (4s vs. 8s, 5s vs. 10s), the resulting outputs do not consistently improve. 
    }
    \label{fig:dur}
    \vspace{-15pt}
\end{figure}
We regard the Chain-of-Frame (CoF) as the thinking process of video reasoning and analyze how video duration influences the correctness of final answers. In CoF, a longer duration corresponds to a longer or more detailed reasoning process, which intuitively might be expected to enhance reasoning accuracy. However, as shown in Fig.~\ref{fig:dur}, our observations reveal a counterintuitive pattern: extending the duration does not consistently lead to better reasoning or higher-quality outputs. Instead, longer sequences often introduce redundant or irrelevant content, and in some cases, cause the model to hallucinate unrelated objects in the final frame.

This phenomenon aligns with findings from prior studies on temporal reasoning, which indicate that increasing sequence length expands the available causal evidence but also magnifies attention drift and temporal mis-binding \cite{Li2024MVBench,Fang2024MMBenchVideo,Zhou2025MLVU,tan2025allvballinonelongvideo}. While longer clips can improve performance when additional frames contain relevant information and the model effectively integrates distant cues, excessive temporal expansion tends to dilute attention and accumulate noise. A deeper investigation into how frame budget, sampling stride, and context window interact with reasoning quality will be an important direction for future work.


\subsection{Video Models vs. Image Models}
\label{subsec:t2i}

\begin{figure*}[ht]
    \centering
    \includegraphics[width=\linewidth]{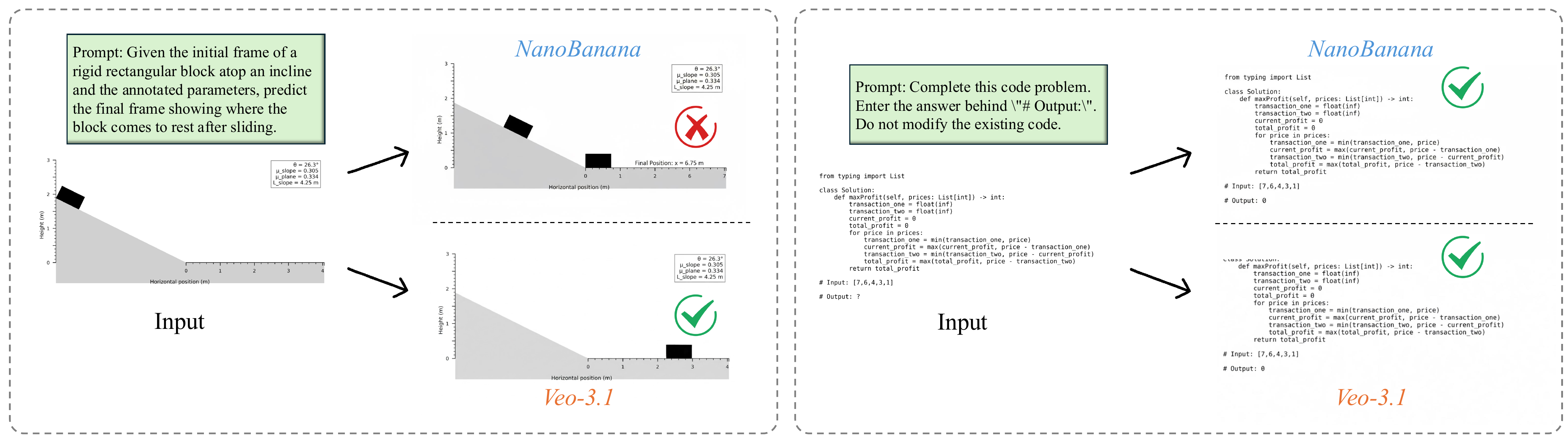}
    \caption{Comparison between Veo-3.1 (video model) and NanoBanana (image model) on \textit{Block Sliding} (left) and \textit{Code Execution} (right). Video models leverage the Chain-of-Frames process to simulate intermediate states, enabling stronger performance on tasks that require causal or temporal reasoning, although intermediate transitions may still appear physically inconsistent. Image models provide clean and stable outputs and excel at text-based tasks such as code execution, but their single-frame reasoning limits their ability to capture the underlying physical dynamics.}
    \label{fig:t2i_comparison}
\end{figure*}

To isolate the contribution of temporal reasoning, we compare Veo-3.1 with NanoBanana~\citep{GoogleDeepMind2025NanoBanana} on V-ReasonBench, treating them as representative video-based and image-based reasoning paradigms. Veo-3.1 is evaluated in full video-generation mode, while NanoBanana functions as a powerful image-only baseline that performs reasoning without temporal progression. Fig.~\ref{fig:t2i_comparison} shows representative outcomes on physics-law and code-reasoning tasks, highlighting the performance gap attributable to temporal modeling.

Image-based models operate on a single static frame and therefore rely heavily on structural priors, textual cues, and pattern recognition. This enables high reliability on code-reasoning and symbolic tasks, where syntax, layout, and character-level precision drive performance. However, the lack of temporal information limits their ability to infer dynamics. When facing tasks involving momentum transfer, balance, collisions, spatial transformations, or chain-structured geometric manipulation, they often choose visually plausible outcomes that do not reflect the correct causal process.

Video models exhibit the opposite strength profile. By generating a CoF sequence, Veo-3.1 can explicitly model transitions, represent latent motion paths, and maintain spatial and causal continuity across time. This frame-wise evolution provides the model with an internal mechanism for simulating physical dynamics and multi-step spatial transformations, which directly improves accuracy on physics-oriented tasks. Importantly, the same CoF mechanism also benefits code-reasoning tasks: intermediate frames act as visual checkpoints that stabilize the symbolic generation process, reducing local inconsistencies and improving step-wise logical execution.

Temporal modeling through CoF gives video models clear advantages in both physical and procedural reasoning. Image models are strong at static structural tasks, while video models leverage process-aware temporal dynamics to handle multi-step, causal, and simulation-heavy problems. Combining precise static parsing with CoF-based temporal modeling offers a promising path toward stronger visual reasoning systems.


\subsection{Hallucination in Video Reasoning}
\begin{figure*}[ht]
    \centering
    \includegraphics[width=\linewidth]{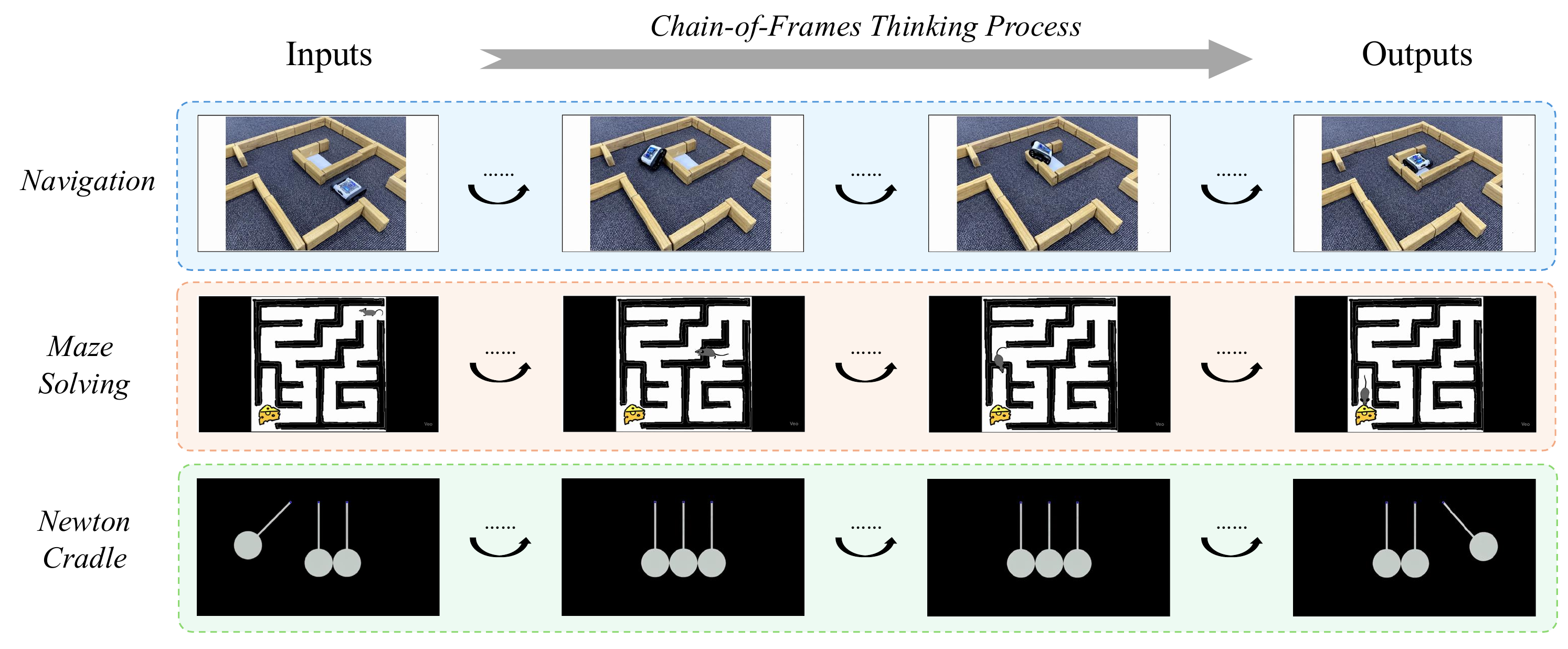}
    \caption{Examples of hallucinations in the Chain-of-Frame reasoning process. Each row shows a trajectory from input to output, where the final frame is correct but intermediate frames display unrealistic or physically inconsistent transitions.}
    \label{fig:hall}
    \vspace{-5pt}
\end{figure*}
During the exploratory stage of benchmark design, we observed several hallucination phenomena emerging within the Chain-of-Frame process. Models may sometimes produce the correct final outcome (last frame) while following an incorrect reasoning process. As illustrated in Fig.~\ref{fig:hall}, in the \textit{maze-solving} task, a mouse successfully reaches the cheese in the final frame even though its intermediate trajectory passes through solid walls. A similar issue also appears in \textit{navigation} tasks. In the \textit{Newton’s cradle} task, the final configuration of moving and stationary balls aligns with the ground-truth label, yet the intermediate frames violate momentum conservation. For instance, when the leftmost ball is released, the entire system remains still instead of transferring motion immediately. These cases exemplify temporal hallucination, where invented or misordered actions and fabricated transitions preserve the correct endpoint but break causal consistency. This phenomenon has been documented in recent evaluations of video language models \cite{Choong2024VidHal,Li2025VidHalluc} and corroborated by multimodal hallucination surveys highlighting their vulnerability to dense or abstract visual patterns \cite{Bai2024MMHSurvey}.

From a benchmarking viewpoint, such “right answer, wrong process” failures are hard to detect if we only check endpoints, and they are also difficult to adjudicate using VLMs as mid-frame judges because VLMs themselves can mis-bind temporal relations or hallucinate missing steps \cite{Choong2024VidHal,Li2025VidHalluc}. 
Accordingly, we favor \emph{end-state–verifiable} tasks where any process error necessarily yields an incorrect terminal state. 

\section{Conclusion}
\label{sec:conclusion}

We present V-ReasonBench, a unified benchmark suite for evaluating video reasoning under the Chain-of-Frame paradigm. Using scalable last-frame scoring across four core reasoning dimensions, it provides a reliable and scalable assessment beyond visual fidelity. Experiments on six state-of-the-art models reveal distinct strengths and systematic failure modes, underscoring the gap between generation quality and true reasoning ability. As an early exploration, V-ReasonBench offers a reproducible foundation for advancing human-aligned video reasoning models.


{
    \small
    \bibliographystyle{ieeenat_fullname}
    \bibliography{main}

@String(CVPR= {IEEE Conf. Comput. Vis. Pattern Recog.})

@String(ICCV= {Int. Conf. Comput. Vis.})

@String(ICLR = {Int. Conf. Learn. Represent.})

@String(AAAI = {AAAI})

@String(CVPR  = {CVPR})

@String(ICCV  = {ICCV})

@String(ICLR  = {ICLR})

@misc{chen2025skyreelsv2infinitelengthfilmgenerative,
      title={SkyReels-V2: Infinite-length Film Generative Model}, 
      author={Guibin Chen and Dixuan Lin and Jiangping Yang and Chunze Lin and Junchen Zhu and Mingyuan Fan and Hao Zhang and Sheng Chen and Zheng Chen and Chengcheng Ma and Weiming Xiong and Wei Wang and Nuo Pang and Kang Kang and Zhiheng Xu and Yuzhe Jin and Yupeng Liang and Yubing Song and Peng Zhao and Boyuan Xu and Di Qiu and Debang Li and Zhengcong Fei and Yang Li and Yahui Zhou},
      year={2025},
      eprint={2504.13074},
      archivePrefix={arXiv},
      primaryClass={cs.CV},
      url={https://arxiv.org/abs/2504.13074}, 
}

@misc{zhang2025show1marryingpixellatent,
      title={Show-1: Marrying Pixel and Latent Diffusion Models for Text-to-Video Generation}, 
      author={David Junhao Zhang and Jay Zhangjie Wu and Jia-Wei Liu and Rui Zhao and Lingmin Ran and Yuchao Gu and Difei Gao and Mike Zheng Shou},
      year={2025},
      eprint={2309.15818},
      archivePrefix={arXiv},
      primaryClass={cs.CV},
      url={https://arxiv.org/abs/2309.15818}, 
}

@misc{ma2025stepvideot2vtechnicalreportpractice,
      title={Step-Video-T2V Technical Report: The Practice, Challenges, and Future of Video Foundation Model}, 
      author={Guoqing Ma and Haoyang Huang and Kun Yan and Liangyu Chen and Nan Duan and Shengming Yin and Changyi Wan and Ranchen Ming and Xiaoniu Song and Xing Chen and Yu Zhou and Deshan Sun and Deyu Zhou and Jian Zhou and Kaijun Tan and Kang An and Mei Chen and Wei Ji and Qiling Wu and Wen Sun and Xin Han and Yanan Wei and Zheng Ge and Aojie Li and Bin Wang and Bizhu Huang and Bo Wang and Brian Li and Changxing Miao and Chen Xu and Chenfei Wu and Chenguang Yu and Dapeng Shi and Dingyuan Hu and Enle Liu and Gang Yu and Ge Yang and Guanzhe Huang and Gulin Yan and Haiyang Feng and Hao Nie and Haonan Jia and Hanpeng Hu and Hanqi Chen and Haolong Yan and Heng Wang and Hongcheng Guo and Huilin Xiong and Huixin Xiong and Jiahao Gong and Jianchang Wu and Jiaoren Wu and Jie Wu and Jie Yang and Jiashuai Liu and Jiashuo Li and Jingyang Zhang and Junjing Guo and Junzhe Lin and Kaixiang Li and Lei Liu and Lei Xia and Liang Zhao and Liguo Tan and Liwen Huang and Liying Shi and Ming Li and Mingliang Li and Muhua Cheng and Na Wang and Qiaohui Chen and Qinglin He and Qiuyan Liang and Quan Sun and Ran Sun and Rui Wang and Shaoliang Pang and Shiliang Yang and Sitong Liu and Siqi Liu and Shuli Gao and Tiancheng Cao and Tianyu Wang and Weipeng Ming and Wenqing He and Xu Zhao and Xuelin Zhang and Xianfang Zeng and Xiaojia Liu and Xuan Yang and Yaqi Dai and Yanbo Yu and Yang Li and Yineng Deng and Yingming Wang and Yilei Wang and Yuanwei Lu and Yu Chen and Yu Luo and Yuchu Luo and Yuhe Yin and Yuheng Feng and Yuxiang Yang and Zecheng Tang and Zekai Zhang and Zidong Yang and Binxing Jiao and Jiansheng Chen and Jing Li and Shuchang Zhou and Xiangyu Zhang and Xinhao Zhang and Yibo Zhu and Heung-Yeung Shum and Daxin Jiang},
      year={2025},
      eprint={2502.10248},
      archivePrefix={arXiv},
      primaryClass={cs.CV},
      url={https://arxiv.org/abs/2502.10248}, 
}

@article{chu2024navigate,
  title={Navigate through Enigmatic Labyrinth: A Survey of Chain‐of-Thought Reasoning in Large Language Models},
  author={Chu, Z and et al.},
  journal={ACL Long 2024},
  year={2024}
}

@misc{kamoi2025visonlyqalargevisionlanguage,
      title={VisOnlyQA: Large Vision Language Models Still Struggle with Visual Perception of Geometric Information}, 
      author={Ryo Kamoi and Yusen Zhang and Sarkar Snigdha Sarathi Das and Ranran Haoran Zhang and Rui Zhang},
      year={2025},
      eprint={2412.00947},
      archivePrefix={arXiv},
      primaryClass={cs.CL},
      url={https://arxiv.org/abs/2412.00947}, 
}

@misc{shinde2025surveyefficientvisionlanguagemodels,
      title={A Survey on Efficient Vision-Language Models}, 
      author={Gaurav Shinde and Anuradha Ravi and Emon Dey and Shadman Sakib and Milind Rampure and Nirmalya Roy},
      year={2025},
      eprint={2504.09724},
      archivePrefix={arXiv},
      primaryClass={cs.CV},
      url={https://arxiv.org/abs/2504.09724}, 
}

@inproceedings{Li2024MVBench,
  author    = {D. Li and Z. Huang and H. Liu and K. Zou and Y. He and F. Zhang and Y. Zhang and J. He and W.-S. Zheng and Y. Qiao and Z. Liu},
  title     = {MVBench: A Comprehensive Multi-Modal Video Understanding Benchmark},
  booktitle = {CVPR},
  year      = {2024},
  arxiv     = {2311.17005},
  note      = {preprint}
}

@inproceedings{Li2025VidHalluc,
  title     = {VidHalluc: Evaluating Temporal Hallucinations in Multimodal Large Language Models},
  author    = {Chong Li and et al.},
  booktitle = {CVPR},
  year      = {2025},
  url       = {https://openaccess.thecvf.com/content/CVPR2025/papers/Li_VidHalluc_Evaluating_Temporal_Hallucinations_in_Multimodal_Large_Language_Models_for_CVPR_2025_paper.pdf}
}

@article{Choong2024VidHal,
  title   = {Benchmarking Temporal Hallucinations in Vision LLMs},
  author  = {W. Y. Choong and et al.},
  journal = {arXiv preprint},
  year    = {2024},
  url     = {https://arxiv.org/abs/2411.16771}
}

@article{Bai2024MMHSurvey,
  title   = {Hallucination of Multimodal Large Language Models},
  author  = {Zhuosheng Bai and et al.},
  journal = {arXiv preprint},
  year    = {2024},
  url     = {https://arxiv.org/abs/2404.18930}
}

@article{Ghazanfari2025CoF,
  title   = {Chain-of-Frames: Advancing Video Understanding in Multimodal LLMs via Frame-Aware Reasoning},
  author  = {S. Ghazanfari and et al.},
  journal = {arXiv preprint arXiv:2506.00318},
  year    = {2025},
  url     = {https://arxiv.org/abs/2506.00318}
}

@inproceedings{Fang2024MMBenchVideo,
  title     = {MMBench-Video: A Long-Form Multi-Shot Benchmark for Video Understanding},
  author    = {Xinyu Fang and et al.},
  booktitle = {NeurIPS Datasets and Benchmarks},
  year      = {2024},
  url       = {https://proceedings.neurips.cc/paper_files/paper/2024/file/a2326c9715a516c91174132e0170073a-Paper-Datasets_and_Benchmarks_Track.pdf}
}

@inproceedings{Zhou2025MLVU,
  title     = {MLVU: Benchmarking Multi-task Long Video Understanding},
  author    = {Jun Zhou and et al.},
  booktitle = {CVPR},
  year      = {2025},
  url       = {https://openaccess.thecvf.com/content/CVPR2025/papers/Zhou_MLVU_Benchmarking_Multi-task_Long_Video_Understanding_CVPR_2025_paper.pdf}
}

@misc{tan2025allvballinonelongvideo,
      title={ALLVB: All-in-One Long Video Understanding Benchmark}, 
      author={Xichen Tan and Yuanjing Luo and Yunfan Ye and Fang Liu and Zhiping Cai},
      year={2025},
      eprint={2503.07298},
      archivePrefix={arXiv},
      primaryClass={cs.CV},
      url={https://arxiv.org/abs/2503.07298}, 
}

@article{Wei2022CoT,
  title   = {Chain-of-Thought Prompting Elicits Reasoning in Large Language Models},
  author  = {Jason Wei and Xuezhi Wang and Dale Schuurmans and Maarten Bosma and Brian Ichter and Fei Xia and Ed H. Chi and Quoc V. Le and Denny Zhou},
  journal = {arXiv preprint arXiv:2201.11903},
  year    = {2022},
  url     = {https://arxiv.org/abs/2201.11903}
}

@inproceedings{Zhou2022LeastToMost,
  title     = {Least-to-Most Prompting Enables Complex Reasoning in Large Language Models},
  author    = {Denny Zhou and Nathanael Schärli and Le Hou and Jason Wei and Nathan Scales and Xuezhi Wang and Dale Schuurmans and Olivier Bousquet and Quoc Le and Ed H. Chi},
  booktitle = {Proceedings of the Advances in Neural Information Processing Systems (NeurIPS)},
  year      = {2022},
  url       = {https://arxiv.org/abs/2205.10625}
}

@article{Wang2022SelfConsistency,
  title   = {Self-Consistency Improves Chain-of-Thought Reasoning in Language Models},
  author  = {Xuezhi Wang and Jason Wei and Dale Schuurmans and Quoc Le and Ed H. Chi and Sharan Narang and Aakanksha Chowdhery and Denny Zhou},
  journal = {arXiv preprint arXiv:2203.11171},
  year    = {2022},
  url     = {https://arxiv.org/abs/2203.11171}
}

@article{wei2022chain,
  title   = {Chain-of-Thought Prompting Elicits Reasoning in Large Language Models},
  author  = {Wei, Jason and Wang, Xuezhi and Schuurmans, Dale and Bosma, Maarten and Ichter, Brian and Xia, Fei and Chi, Ed and Le, Quoc and Zhou, Denny},
  journal = {Advances in Neural Information Processing Systems},
  volume  = {35},
  pages   = {24824--24837},
  year    = {2022}
}

@inproceedings{Yi2020CLEVRER,
  title     = {CLEVRER: Collision Events for Video Representation and Reasoning},
  author    = {Yi, Kexin and Gan, Chuang and Li, Yunzhu and Kohli, Pushmeet and Wu, Jiajun and Tenenbaum, Joshua B. and Torralba, Antonio},
  booktitle = {ICLR},
  year      = {2020}
}

@article{Girdhar2019CATER,
  title   = {CATER: A Diagnostic Dataset for Compositional Actions and Temporal Reasoning},
  author  = {Girdhar, Rohit and Ramanan, Deva},
  year    = {2019},
  journal = {arXiv preprint arXiv:1910.04744}
}

@inproceedings{GrundeMcLaughlin2021AGQA,
  title     = {AGQA: A Benchmark for Compositional Spatio-Temporal Reasoning},
  author    = {Gr{\"u}nde-McLaughlin, Michaela and Krishna, Ranjay and Niebles, Juan Carlos and Fei-Fei, Li},
  booktitle = {CVPR},
  year      = {2021}
}

@misc{comanici2025gemini25pushingfrontier,
      title={Gemini 2.5: Pushing the Frontier with Advanced Reasoning, Multimodality, Long Context, and Next Generation Agentic Capabilities}, 
      author={Gheorghe Comanici and Eric Bieber and Mike Schaekermann and Ice Pasupat and Noveen Sachdeva and Inderjit Dhillon and Marcel Blistein and Ori Ram and Dan Zhang and Evan Rosen and Luke Marris and Sam Petulla and et al},
      year={2025},
      eprint={2507.06261},
      archivePrefix={arXiv},
      primaryClass={cs.CL},
      url={https://arxiv.org/abs/2507.06261}, 
}

@inproceedings{Pothiraj2025CAPTURE,
  title     = {CAPTURe: Evaluating Spatial Reasoning in Vision Language Models via Occluded Object Counting},
  author    = {Atin Pothiraj and Elias Stengel-Eskin and Jaemin Cho and Mohit Bansal},
  booktitle = {ICCV},
  year      = {2025},
  url       = {https://openaccess.thecvf.com/content/ICCV2025/papers/Pothiraj_CAPTURE_Evaluating_Spatial_Reasoning_in_Vision_Language_Models_via_Occluded_ICCV_2025_paper.pdf}
}

@inproceedings{Li2024TopViewSpatial,
  title     = {Vision-Language Models as Top-View Spatial Reasoners},
  author    = {Chengxi Li and Wenxuan Zhang and Xuwu Sun and Zhaoxiang Zhang and Yuntao Chen and Kaipeng Zhang},
  booktitle = {EMNLP},
  year      = {2024},
  url       = {https://aclanthology.org/2024.emnlp-main.106.pdf}
}

@article{Wiedemer2025VideoZeroShot,
  title        = {Video Models Are Zero-Shot Learners and Reasoners},
  author       = {Thaddäus Wiedemer and Yuxuan Li and Paul Vicol and Shixiang Shane Gu and Nick Matarese and Kevin Swersky and Been Kim and Priyank Jaini and Robert Geirhos},
  journal      = {arXiv preprint arXiv:2509.20328},
  year         = {2025},
  url          = {https://arxiv.org/abs/2509.20328}
}

@misc{kuaishou2024kling,
  title   = {Kling AI: Next-Generation AI Creative Studio},
  author  = {{Kuaishou Technology}},
  year    = {2024},
  month   = {June},
  howpublished = {\url{https://klingai.com/}}
}

@techreport{openai2025sora2,
  title       = {Sora 2 System Card},
  author      = {{OpenAI}},
  institution = {OpenAI},
  year        = {2025},
  month       = {September},
  url         = {https://cdn.openai.com/pdf/50d5973c-c4ff-4c2d-986f-c72b5d0ff069/sora_2_system_card.pdf}
}

@article{gao2025seedance,
  title={Seedance 1.0: Exploring the Boundaries of Video Generation Models},
  author={Gao, Yu and Guo, Haoyuan and Hoang, Tuyen and Huang, Weilin and Jiang, Lu and Kong, Fangyuan and Li, Huixia and Li, Jiashi and Li, Liang and Li, Xiaojie and others},
  journal={arXiv preprint arXiv:2506.09113},
  year={2025}
}

@techreport{minimax2025hailuo02,
  title        = {Hailuo 02 Model Card},
  author       = {{MiniMax AI}},
  institution  = {MiniMax AI},
  year         = {2025},
  month        = {June},
  url          = {https://hailuo-02.com/}
}

@techreport{shengshu2025viduq2,
  title        = {Vidu Q2 Model Release},
  author       = {{ShengShu Technology}},
  institution  = {ShengShu Technology},
  year         = {2025},
  month        = {October},
  url          = {https://www.scmp.com/tech/tech-trends/article/3329800/chinese-ai-start-shengshu-unveils-vidu-q2-challenge-openais-sora}
}

@misc{ravi2024sam2segmentimages,
      title={SAM 2: Segment Anything in Images and Videos}, 
      author={Nikhila Ravi and Valentin Gabeur and Yuan-Ting Hu and Ronghang Hu and Chaitanya Ryali and Tengyu Ma and Haitham Khedr and Roman Rädle and Chloe Rolland and Laura Gustafson and Eric Mintun and Junting Pan and Kalyan Vasudev Alwala and Nicolas Carion and Chao-Yuan Wu and Ross Girshick and Piotr Dollár and Christoph Feichtenhofer},
      year={2024},
      eprint={2408.00714},
      archivePrefix={arXiv},
      primaryClass={cs.CV},
      url={https://arxiv.org/abs/2408.00714}, 
}

@misc{runway2024gen3,
  title        = {Introducing Gen-3 Alpha: A New Generation of Generative Video Models},
  author       = {{Runway Research}},
  year         = {2024},
  howpublished = {\url{https://runwayml.com/research/introducing-gen-3-alpha}},
}

@misc{google2025veo3,
  title        = {Veo 3.1: Google DeepMind Video Generation Model},
  author       = {{Google DeepMind}},
  year         = {2025},
  howpublished = {\url{https://deepmind.google/models/veo/}},
}

@misc{tencent2024hunyuanvideo,
  title        = {HunyuanVideo: A Systematic Framework for Large Video Generation Models},
  author       = {{Tencent Hunyuan Video Team}},
  year         = {2024},
  howpublished = {\url{https://github.com/Tencent-Hunyuan/HunyuanVideo}},
}

@article{wan2025wan,
  title   = {Wan: Open and Advanced Large-Scale Video Generative Models},
  author  = {Wang, Ang and Ai, Baole and Wen, Bin and Mao, Chaojie and Xie, Chen-Wei and Chen, Di and Yu, Feiwu and Zhao, Haiming and Yang, Jianxiao and others},
  journal = {arXiv preprint arXiv:2503.20314},
  year    = {2025},
}

@misc{chen2024gentrondiffusiontransformersimage,
      title={GenTron: Diffusion Transformers for Image and Video Generation}, 
      author={Shoufa Chen and Mengmeng Xu and Jiawei Ren and Yuren Cong and Sen He and Yanping Xie and Animesh Sinha and Ping Luo and Tao Xiang and Juan-Manuel Perez-Rua},
      year={2024},
      eprint={2312.04557},
      archivePrefix={arXiv},
      primaryClass={cs.CV},
      url={https://arxiv.org/abs/2312.04557}, 
}

@misc{ma2024lattelatentdiffusiontransformer,
      title={Latte: Latent Diffusion Transformer for Video Generation}, 
      author={Xin Ma and Yaohui Wang and Gengyun Jia and Xinyuan Chen and Ziwei Liu and Yuan-Fang Li and Cunjian Chen and Yu Qiao},
      year={2024},
      eprint={2401.03048},
      archivePrefix={arXiv},
      primaryClass={cs.CV},
      url={https://arxiv.org/abs/2401.03048}, 
}

@misc{zheng2024opensorademocratizingefficientvideo,
      title={Open-Sora: Democratizing Efficient Video Production for All}, 
      author={Zangwei Zheng and Xiangyu Peng and Tianji Yang and Chenhui Shen and Shenggui Li and Hongxin Liu and Yukun Zhou and Tianyi Li and Yang You},
      year={2024},
      eprint={2412.20404},
      archivePrefix={arXiv},
      primaryClass={cs.CV},
      url={https://arxiv.org/abs/2412.20404}, 
}

@misc{gao2024luminat2xtransformingtextmodality,
      title={Lumina-T2X: Transforming Text into Any Modality, Resolution, and Duration via Flow-based Large Diffusion Transformers}, 
      author={Peng Gao and Le Zhuo and Dongyang Liu and Ruoyi Du and Xu Luo and Longtian Qiu and Yuhang Zhang and Chen Lin and Rongjie Huang and Shijie Geng and Renrui Zhang and Junlin Xi and Wenqi Shao and Zhengkai Jiang and Tianshuo Yang and Weicai Ye and He Tong and Jingwen He and Yu Qiao and Hongsheng Li},
      year={2024},
      eprint={2405.05945},
      archivePrefix={arXiv},
      primaryClass={cs.CV},
      url={https://arxiv.org/abs/2405.05945}, 
}

@misc{yang2025cogvideoxtexttovideodiffusionmodels,
      title={CogVideoX: Text-to-Video Diffusion Models with An Expert Transformer}, 
      author={Zhuoyi Yang and Jiayan Teng and Wendi Zheng and Ming Ding and Shiyu Huang and Jiazheng Xu and Yuanming Yang and Wenyi Hong and Xiaohan Zhang and Guanyu Feng and Da Yin and Yuxuan Zhang and Weihan Wang and Yean Cheng and Bin Xu and Xiaotao Gu and Yuxiao Dong and Jie Tang},
      year={2025},
      eprint={2408.06072},
      archivePrefix={arXiv},
      primaryClass={cs.CV},
      url={https://arxiv.org/abs/2408.06072}, 
}

@article{huang2025vchain,
  title={VChain: Chain-of-Visual-Thought for Reasoning in Video Generation},
  author={Huang, Ziqi and others},
  journal={arXiv preprint arXiv:2510.05094},
  year={2025}
}

@inproceedings{chen2024visual,
  title={Visual Chain-of-Thought Prompting for Knowledge-Based Visual Reasoning},
  author={Chen, Zhuowan and Zhou, Qihang and Shen, Yibo and Hong, Yuzhe and Sun, Zhiyuan and Gutfreund, Dan and Gan, Chuang},
  booktitle={Proceedings of the AAAI Conference on Artificial Intelligence},
  volume={38},
  number={2},
  pages={1244--1252},
  year={2024}
}

@article{zhao2025cotvla,
  title={CoT-VLA: Visual Chain-of-Thought Reasoning for Vision-Language-Action Models},
  author={Zhao, Xinyi and others},
  journal={arXiv preprint arXiv:2503.22020},
  year={2025}
}

@misc{wang2025videorftincentivizingvideoreasoning,
      title={VideoRFT: Incentivizing Video Reasoning Capability in MLLMs via Reinforced Fine-Tuning}, 
      author={Qi Wang and Yanrui Yu and Ye Yuan and Rui Mao and Tianfei Zhou},
      year={2025},
      eprint={2505.12434},
      archivePrefix={arXiv},
      primaryClass={cs.CV},
      url={https://arxiv.org/abs/2505.12434}, 
}

@misc{li2025videochatr1enhancingspatiotemporalperception,
      title={VideoChat-R1: Enhancing Spatio-Temporal Perception via Reinforcement Fine-Tuning}, 
      author={Xinhao Li and Ziang Yan and Desen Meng and Lu Dong and Xiangyu Zeng and Yinan He and Yali Wang and Yu Qiao and Yi Wang and Limin Wang},
      year={2025},
      eprint={2504.06958},
      archivePrefix={arXiv},
      primaryClass={cs.CV},
      url={https://arxiv.org/abs/2504.06958}, 
}

@article{HaCohen2024LTXVideo,
  title={LTX-Video: Realtime Video Latent Diffusion},
  author={HaCohen, Yoav and Chiprut, Nisan and Brazowski, Benny and Shalem, Daniel and Moshe, Dudu and Richardson, Eitan and Levin, Eran and Shiran, Guy and Zabari, Nir and Gordon, Ori and Panet, Poriya and Weissbuch, Sapir and Kulikov, Victor and Bitterman, Yaki and Melumian, Zeev and Bibi, Ofir},
  journal={arXiv preprint arXiv:2501.00103},
  year={2024}
}

@misc{polyak2025moviegencastmedia,
      title={Movie Gen: A Cast of Media Foundation Models}, 
      author={Adam Polyak and Amit Zohar and Andrew Brown and Andros Tjandra and Animesh Sinha and Ann Lee and Apoorv Vyas and Bowen Shi, et al.},
      year={2025},
      eprint={2410.13720},
      archivePrefix={arXiv},
      primaryClass={cs.CV},
      url={https://arxiv.org/abs/2410.13720}, 
}

@misc{GoogleDeepMind2025NanoBanana,
  author       = {Fortin, Alisa and Vernade, Guillaume and Kampf, Kat and Reshi, Ammaar},
  title        = {Introducing {Gemini 2.5 Flash Image} (aka “Nano Banana”)},
  howpublished = {Blog post, Google Developers},
  year         = {2025},
  month        = {Aug},
  note         = {Available at: https://developers.googleblog.com/en/introducing-gemini-2-5-flash-image/},
}
}

\clearpage


\onecolumn

\section{Additional Details for Each Task}
\label{appendix:details}

\subsection{Structured Problem-Solving}
This reasoning class targets systematic computational and logical reasoning abilities. The four tasks collectively assess models' capacity to handle rule-driven problems that require step-by-step execution, numerical manipulation, and strategic thinking.

\begin{figure}[!h]
    \centering
    \includegraphics[width=\linewidth]{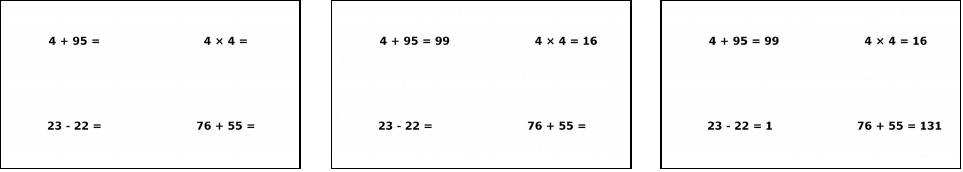}
    \caption{An illustration of the arithmetic operation task, showing the starting frame (left), an intermediate frame (middle), and the final frame with completed answers (right).}
    \label{fig:arithmetic}
\end{figure}

\paragraph{Arithmetic Operation.} 
Models compute mathematical expressions presented visually, including addition, subtraction, multiplication, and division. Problems are categorized by difficulty: easy (addition/subtraction, range 1-15), medium (addition/subtraction/multiplication, range 1-50), and hard (all four operations, range 1-100). Each instance may contain 2 or 4 problems. All instances are programmatically generated with randomized numerical values and operator combinations, with visual representations using standard mathematical notation rendered in various fonts and sizes to ensure diversity. Fig.~\ref{fig:arithmetic} illustrates an example of the generated video, showing the starting frame with the math problems (left), an intermediate frame (middle), and the final frame with all answers filled in (right). The evaluation prompt is: \textit{``Complete the math problems in the picture. Enter the final answer directly. Do not change anything else in the picture. Static camera, no zoom, no pan, no dolly.''} We employ VLM-based evaluation where Gemini-2.5-Pro extracts all numerical results from the final frame. The prediction is marked as \textbf{passed} only if all problems and their answers are completely correct.

\begin{figure}[!h]
    \centering
    \includegraphics[width=\linewidth]{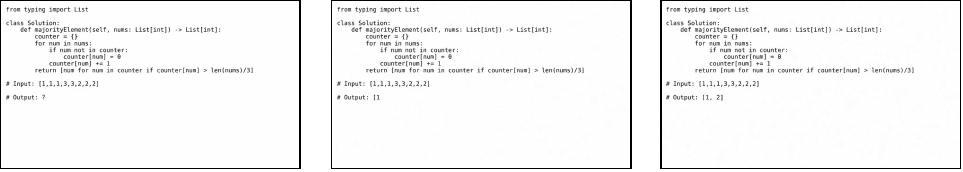}
    \caption{An illustration of the code execution task, showing the starting frame with code and input (left), an intermediate frame (middle), and the final frame with the output filled in (right).}
    \label{fig:code_execution}
\end{figure}

\paragraph{Code Execution.}
Models simulate program execution by tracing through visualized Python code with given inputs to predict outputs. The task includes three difficulty levels: easy, medium, and hard, based on the complexity of the Python code problems sourced from LeetCode. Each instance contains only 1 problem. Code snippets are curated from classical algorithm problems, with inputs and expected outputs synthetically generated. Fig.~\ref{fig:code_execution} illustrates an example of the generated video, showing the starting frame with the code and input (left), an intermediate frame (middle), and the final frame with the output correctly filled in (right). The evaluation prompt is: \textit{``Complete this code problem. Enter the answer behind ``\# Output:''. Do not modify the existing code.''} VLM-based evaluation extracts the code problem and output from the final frame. The prediction \textbf{passes} only if the extracted problem and result exactly match the ground truth.

\begin{figure}[!h]
    \centering
    \includegraphics[width=\linewidth]{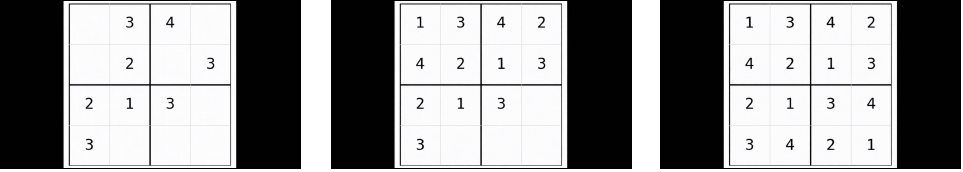}
    \caption{An illustration of the $4\times 4$ Sudoku task, showing the starting frame with the initial puzzle (left), an intermediate solving frame (middle), and the final frame with the completed solution (right).}
    \label{fig:sudoku}
\end{figure}

\paragraph{Sudoku.}
Models complete partially filled $4 \times 4$ or $9\times 9$ Sudoku grids, with difficulty levels determined by the number of pre-filled cells: easy (more givens), medium (moderate givens), hard (fewer givens requiring deeper logical deduction). Puzzles are generated using constraint satisfaction algorithms that ensure unique solutions, and each puzzle is validated for solvability before inclusion with visual grids using clear cell boundaries and digit rendering. Fig.~\ref{fig:sudoku} illustrates an example of the generated video, showing the starting frame with the initial puzzle (left), an intermediate solving stage (middle), and the final frame with the complete solution (right). The evaluation prompt is: \textit{``Create a static, smooth, animation that solves the given [4$\times$4 or 9$\times$9] sudoku. Enter the missing numbers one by one. Do not change anything else in the picture. Only fill the numbers in the empty cells so the sudoku is solved properly. A cursor moves and fills the correct number in the empty boxes.''} (The grid size in the prompt matches the actual puzzle size.) VLM-based evaluation extracts all cell values from the last frame, and a puzzle \textbf{passes} only if all cells are the same numbers as the ground truth.

\begin{figure}[!h]
    \centering
    \includegraphics[width=.8\linewidth]{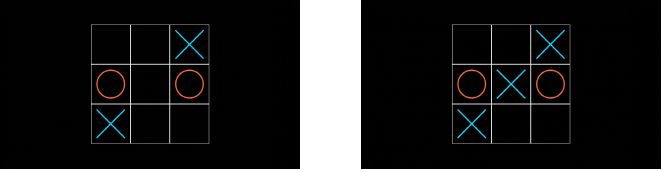}
    \caption{An illustration of the Tic-Tac-Toe task, showing the starting game state (left) and the final frame with the optimal move placed (right).}
    \label{fig:tictactoe}
\end{figure}

\paragraph{Tic-Tac-Toe.}
Models identify optimal moves in $3\times3$ Tic-Tac-Toe games across different scenarios: 2 `X'/`O', 3 `X'/`O', and 4 `X'/`O'. Game states are programmatically generated to cover diverse tactical situations, with each state validated to have a clearly optimal move according to the minimax strategy. Fig.~\ref{fig:tictactoe} illustrates an example of the generated video, showing the starting game state (left) and the final frame with the optimal move placed (right). The evaluation prompt is: \textit{``Given the current state of the Tic-Tac-Toe board in the image, predict the next move that results in a win for the player whose turn it is. Output the position of the winning move.''} Grid-based evaluation partitions the board into individual cells and measures accuracy by comparing each predicted cell with its ground-truth counterpart; a prediction is marked as \textbf{passed} only when all cells match exactly.

\subsection{Spatial Cognition}
This category focuses on visual-spatial reasoning, testing how well models comprehend geometric properties, spatial transformations, and positional relationships in 2D environments.

\begin{figure}[!h]
    \centering
    \includegraphics[width=\linewidth]{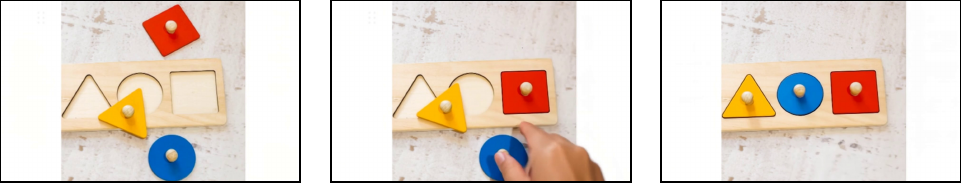}
    \caption{An illustration of the shape fitting task, showing the starting frame with pieces and empty holes (left), an intermediate fitting process (middle), and the final frame with all pieces correctly fitted (right).}
    \label{fig:shape_fitting}
\end{figure}

\paragraph{Shape Fitting.}
Models receive a collection of geometric pieces and a target outline, then must determine the correct arrangement through rotation and positioning, with difficulty varying by shape complexity (simple triangles/rectangles vs. irregular polygons) and the number of pieces (2-5 pieces). 
The images were collected from the Internet or generated by image generation models. Fig.~\ref{fig:shape_fitting} illustrates an example of the generated video, showing the starting frame with separate pieces and empty holes (left), an intermediate fitting process (middle), and the final frame with all pieces correctly placed in their corresponding holes (right).
The evaluation prompt is: \textit{``The scene shows three colored pieces, and a panel with three holes. Each colored piece fits into one and only one hole. A hand grabs each colored piece and puts it into an empty hole that has the exact same shape - if it doesn't fit, the hand tries another hole. All the objects must be placed in their respective holes.''} 
VLM-based evaluation uses Gemini-2.5-Pro~\citep{comanici2025gemini25pushingfrontier} to assess whether all pieces are correctly positioned. A trial is considered a \textbf{pass} only if the VLM confirms that all pieces are properly fitted without any shape alteration.

\begin{figure}[!h]
    \centering
    \includegraphics[width=\linewidth]{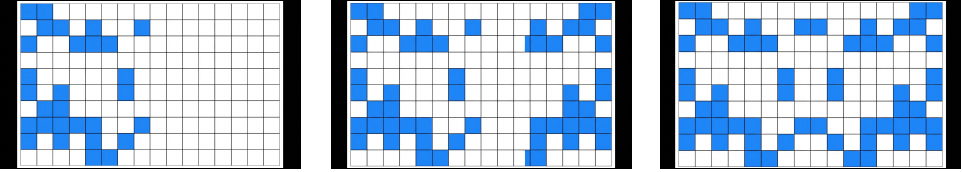}
    \caption{An illustration of the vertical symmetric in the visual symmetry task, showing the starting frame with an incomplete pattern (left), an intermediate completion stage (middle), and the final frame with the fully symmetric pattern (right).}
    \label{fig:symmetry}
\end{figure}

\paragraph{Visual Symmetry.}
Models complete incomplete symmetric patterns, with tasks categorized by symmetry type: reflective (mirror across vertical/horizontal/diagonal axes) and $180^\circ$ rotational symmetry. For diagonal symmetry, patterns are generated on $8\times8$ grids by randomly coloring cells; for other symmetry types, $10\times16$ grids are used with the same random coloring process. Partial patterns are created by showing only half of the symmetric structure, requiring models to infer and complete the missing portion. Fig.~\ref{fig:symmetry} illustrates an example of the generated video, showing the starting frame with the incomplete pattern (left), an intermediate completion stage (middle), and the final frame with the fully symmetric pattern (right). The evaluation prompt is: \textit{``Instantly reflect this pattern [along the central, vertical axis or along the central, horizontal axis or along the main diagonal axis (top-left to bottom-right) or $180^\circ$ along the central, vertical axis] while keeping the existing colored pattern without modification. Static camera perspective, no zoom or pan.''} Grid-based evaluation divides the image into cells and compares cell-wise matching between predicted and ground truth patterns, and a prediction \textbf{passes} only if all cells match exactly.

\begin{figure}[!h]
    \centering
    \includegraphics[width=\linewidth]{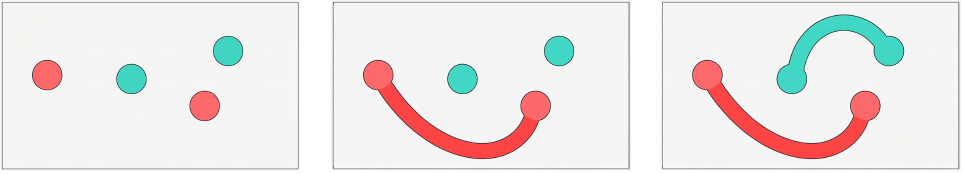}
    \caption{An illustration of the color connection task, showing the starting frame with colored endpoints (left), an intermediate drawing stage (middle), and the final frame with all paths correctly connected (right).}
    \label{fig:color_connection}
\end{figure}

\paragraph{Color Connection.}
Models link colored endpoints through non-intersecting paths on grids, with difficulty varying by the number of color pairs (2-4 pairs). Puzzles are generated with random positions for all circles. Fig.~\ref{fig:color_connection} illustrates an example of the generated video, showing the starting frame with colored endpoints (left), an intermediate path-drawing stage (middle), and the final frame with all color pairs correctly connected (right). The evaluation prompt is: \textit{``Draw three curves, one connecting each pair of circles of the same color. Do not change anything else in the picture. Static camera, no zoom, no pan, no dolly.''} VLM-based evaluation assesses path connectivity and non-intersection, and the prediction \textbf{passes} if the VLM confirms all color pairs are correctly connected without crossings.

\subsection{Pattern-based Inference}
This reasoning class examines inductive capabilities and abstract thinking, focusing on how models extract implicit rules from limited observations and generalize them to new situations.

\begin{figure}[!h]
    \centering
    \includegraphics[width=\linewidth]{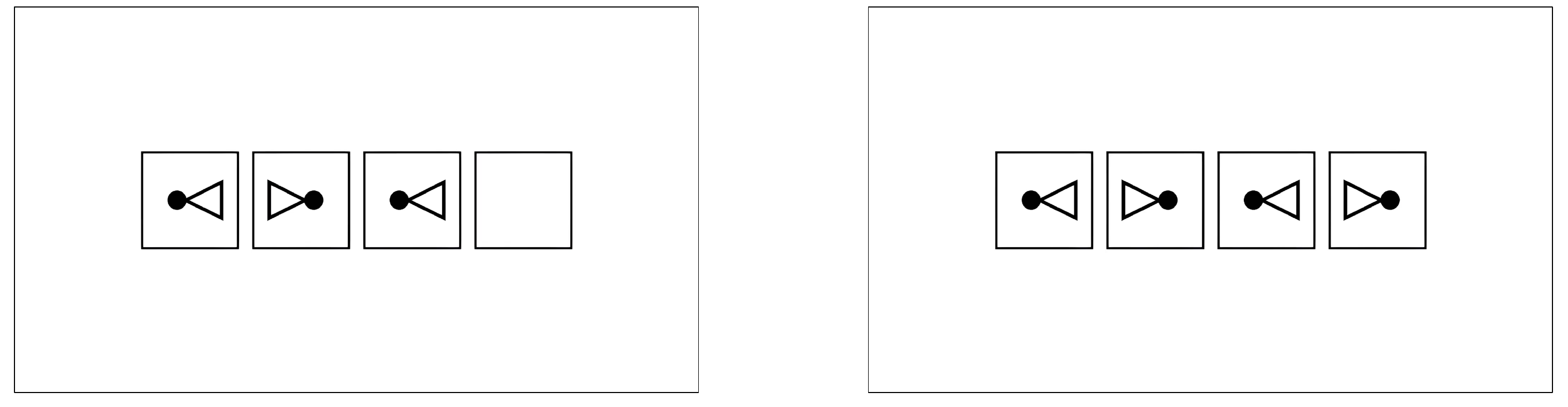}
    \caption{An illustration of the sequence completion task, showing the starting frame with one blank cell (left) and the final frame with all completed cells (right).}
    \label{fig:sequence}
\end{figure}

\paragraph{Sequence Completion.}
Models infer the next element in a visual sequence by recognizing how objects evolve across successive frames. Each sequence introduces variation in object count, shape, and color, requiring the model to detect consistent visual regularities and extend them to the next plausible state. Sequences are programmatically generated using rule-based transformations, with each sequence following a deterministic rule that governs element transitions, and 3-5 examples are provided before asking for a prediction. Fig.~\ref{fig:sequence} shows one demonstration of this task. The evaluation prompt is: \textit{``Finish the incomplete frames of this shape-sequence puzzle. Do not change given patterns. Each row contains equal-sized square cells outlined in black on white. Shapes are solid black arrows, dots, or stripes that obey a clear mathematical rule (rotation, translation, scaling, or toggling). Produce the missing panels so the sequence remains consistent and has a single logical solution.''} Mask-based evaluation compares the predicted next element against ground truth using pixel-level accuracy within the target region, and a prediction \textbf{passes} if Acc $>$ 0.90.

\begin{figure}[!h]
    \centering
    \includegraphics[width=.8\linewidth]{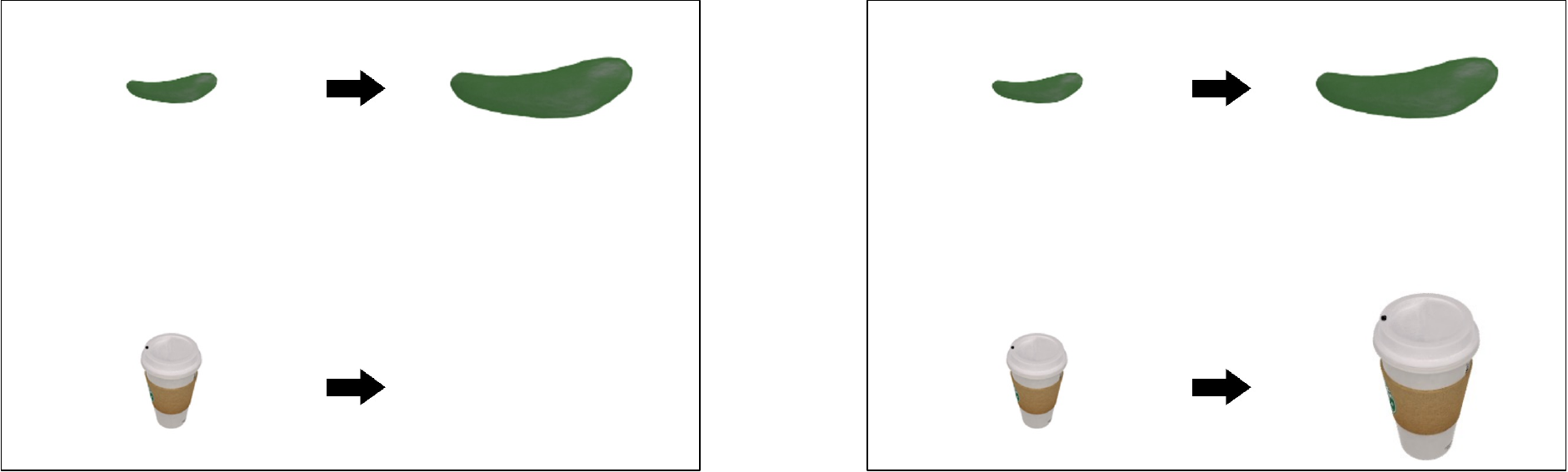}
    \caption{An illustration of the scaling relation in the analogy-solving task, showing the starting frame with the exemplar transformation in the top row (left), and the final frame with the corresponding transformation applied to the target object in the bottom row (right).}
    \label{fig:analogy}
\end{figure}

\paragraph{Analogy Solving.}
Models solve visual analogies of the form ``A:B::C:?", testing different relationship types: spatial transformations (rotation, reflection, scaling), attribute changes (color, texture), and structural modifications (adding/removing components). Analogy problems are created by first defining a transformation, applying it to create the A→B pair, then providing a different C and asking models to apply the same transformation. Fig.~\ref{fig:analogy} presents an example of this relationship, showing the A:B:C triplet and ground truth answer. The evaluation prompt is:
\textit{``Create a smooth animation to generate the missing object in the lower right region and solve the visual analogy. The original three objects must remain still. Static shot, no zoom no pan no dolly.``}
Mask-based evaluation measures pixel-wise similarity between predicted and ground truth answer images at the expected location for the transformed object, and a prediction \textbf{passes} if the mean MSE distance $<$ 0.1.

\begin{figure}[!h]
    \centering
    \includegraphics[width=.8\linewidth]{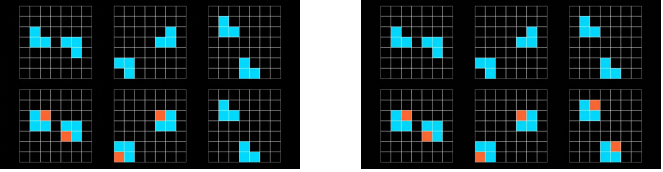}
    \caption{An illustration of the rule following task, showing the starting frame with demonstration pairs and empty test grid (left) and the final frame with the rule correctly applied to the test grid (right).}
    \label{fig:rule_following}
\end{figure}

\paragraph{Rule Following.}
Models infer transformation rules from example pairs and apply them to novel inputs, with rules varying in grid size  and number of colored components in each grid. Rule sets are synthetically defined, with 2-4 demonstration pairs generated per rule, and test cases use different inputs but follow the same underlying transformation logic. Fig.~\ref{fig:rule_following} illustrates an example of the generated video, showing the starting frame with demonstration pairs and an empty test grid (left) and the final frame with the rule correctly applied to the test grid (right). The evaluation prompt is: \textit{``Modify the lower-right grid to adhere to the rule established by the other grids. You can fill cells, clear cells, or change a cell’s color in the lower-right grid only. Don’t modify any of the other grids. Static scene, no zoom, no pan, no dolly.''} Grid-based evaluation partitions the image into grids and verifies the cell-wise correctness of the bottom-right table to be predicted against the ground truth; a prediction \textbf{passes} only when all cells match perfectly (100\% accuracy).

\subsection{Physical Dynamics}
This category assesses intuitive physics understanding, probing whether models can predict real-world physical behaviors based on visual observations of forces, materials, and environmental conditions.

\begin{figure}[!h]
    \centering
    \includegraphics[width=\linewidth]{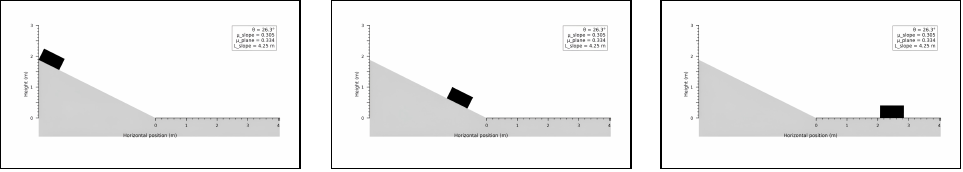}
    \caption{An illustration of the block sliding task, showing the starting frame with objects on an inclined surface (left), an intermediate sliding motion (middle), and the final frame showing which objects slid down (right).}
    \label{fig:block_sliding}
\end{figure}

\paragraph{Block Sliding.}
Models predict whether objects will slide down inclined planes, with scenarios varying by slope angle (15°-60°), object shape (cube, sphere, cylinder), and surface properties (smooth, rough). Scenes are programmatically generated through controlled code specifications, and each scenario is verified to ensure a single, unambiguous outcome regarding whether sliding should occur under the defined conditions. Fig.~\ref{fig:block_sliding} illustrates an example of the generated video, showing the starting frame with objects on the inclined surface (left), an intermediate sliding motion (middle), and the final frame showing the final positions of all objects (right). The evaluation prompt is: \textit{``Observe the objects on the inclined surface. Predict which objects will slide down and which will remain stationary. Show the final state after gravity acts.''} Mask-based evaluation compares object positions in the final frame against ground truth, checking if each object's position indicates sliding or staying, and a prediction \textbf{passes} if all object states are correctly predicted.

\begin{figure}[!h]
    \centering
    \includegraphics[width=\linewidth]{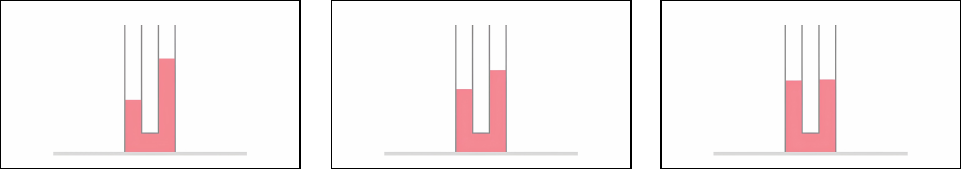}
    \caption{An illustration of the communicating vessels task, showing the starting frame with initial liquid levels (left), an intermediate redistribution stage (middle), and the final frame at equilibrium (right).}
    \label{fig:communicating_vessels}
\end{figure}

\paragraph{Communicating Vessels.}
Models predict liquid level redistribution in communicating vessels of diverse shapes and liquid colors. The simulations compute final states based on hydrostatic equilibrium to ensure physical accuracy. As shown in Fig.~\ref{fig:communicating_vessels} illustrates an example of the generated video, showing the starting frame with initial liquid levels (left), an intermediate redistribution process (middle), and the final frame at equilibrium (right). The evaluation prompt, \textit{``A connected container (communicating vessel) with two vertical tubes filled with liquid. The liquid level in both tubes changes slowly and smoothly until it reaches equilibrium, where the levels remain still. The motion should appear smooth and continuous, with no splashing or turbulence"}, guides the assessment. Performance is measured using a mask-based method that computes the height difference between predicted and ground truth levels in each vessel. A prediction is considered \textbf{passed} if the equilibrated liquid positions are level.

\begin{figure}[!h]
    \centering
    \includegraphics[width=\linewidth]{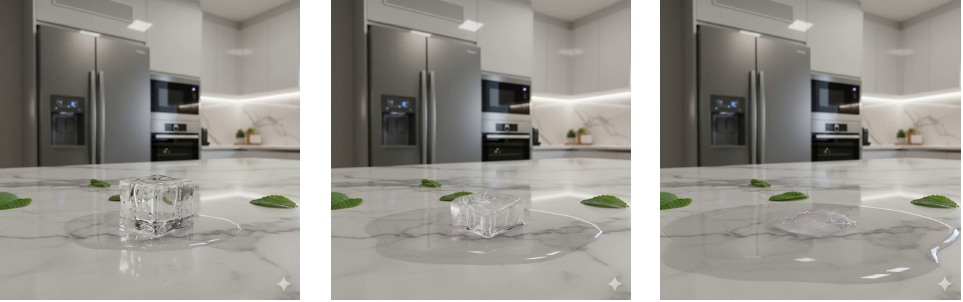}
    \caption{An illustration of the temperature-induced deformation task using an ice melting example, showing the starting frame with solid ice (left), an intermediate melting stage (middle), and the final frame with completely melted ice/water (right).}
    \label{fig:temperature_deformation}
\end{figure}

\paragraph{Temperature-Induced Deformation.}
Models predict material (ice) changes under temperature variations across scenarios, with temperature ranges varying from -20°C to 100°C. Material transformations are created using physics-based rendering and simulation, with each scenario specifying specific temperature and material properties to ensure deterministic outcomes. Fig.~\ref{fig:temperature_deformation} illustrates an example of the generated video showing ice melting, with the starting frame showing solid ice at low temperature (left), an intermediate melting stage (middle), and the final frame showing the completely melted state (right). The evaluation prompt is: \textit{``Observe the material at the initial temperature. Predict how it will change (melt, expand, contract, deform) when the temperature changes to the target value. Show the final state.''} VLM-based evaluation uses Gemini-2.5-Pro~\citep{comanici2025gemini25pushingfrontier} to assess whether the predicted material state matches expected physical behavior (e.g., ice should melt above 0°C, metal should expand when heated), and a prediction \textbf{passes} if the VLM confirms correct physical transformation.

\section{Complete Task-Level Results}

\begin{figure}[!h]
    \centering
    \includegraphics[width=\linewidth]{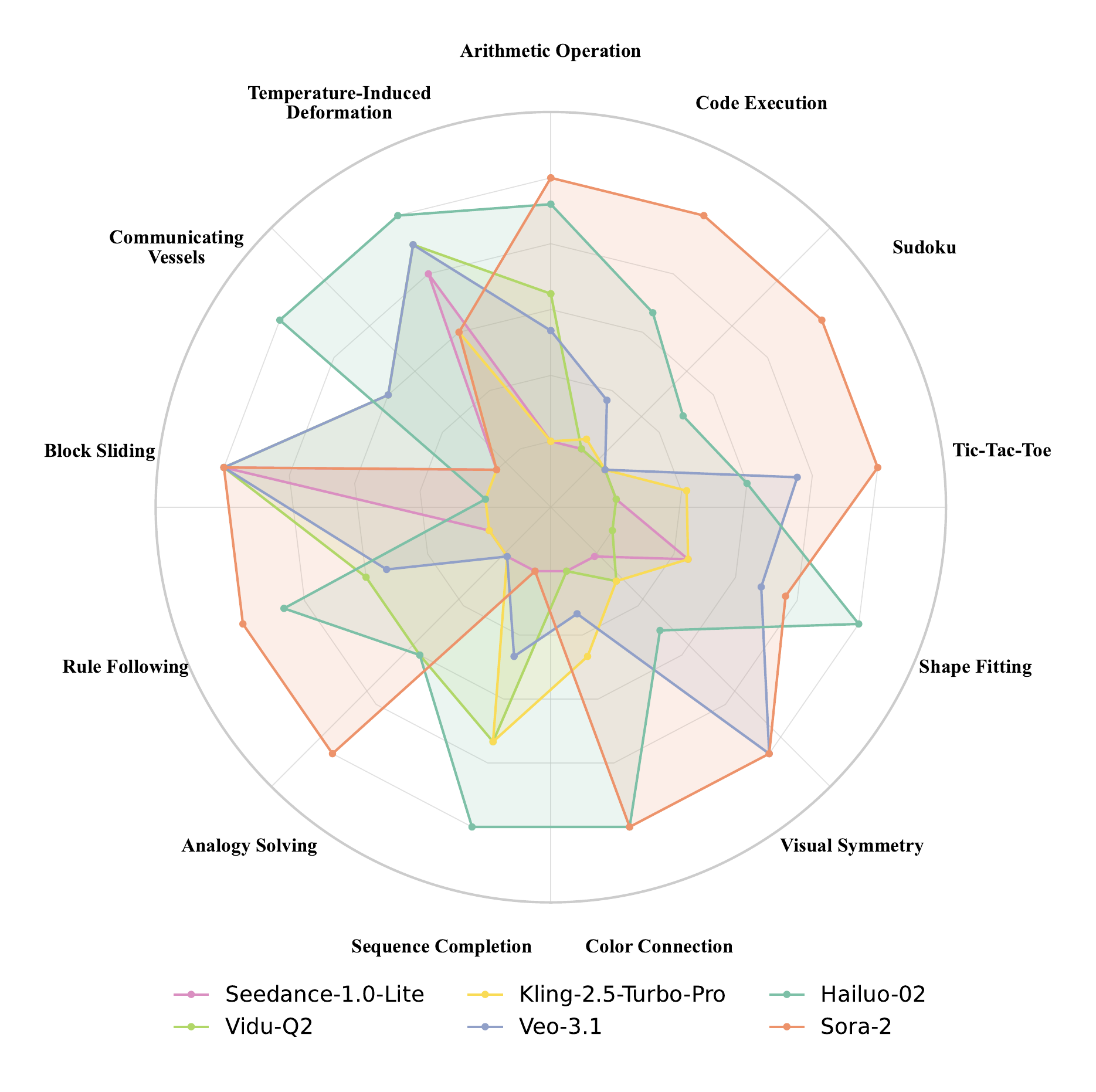}
    \caption{Summary of V-ReasonBench performance across video models. The figure illustrates how six video generation models perform on 13 reasoning tasks, with scores rescaled within each dimension to enable direct comparison.}
    \label{fig:task_visual}
\end{figure}

Tab.~\ref{tab:complete_results} and Fig.~\ref{fig:task_visual} present the detailed pass@5 performance of all evaluated models across each of the 13 reasoning tasks in V-ReasonBench. This granular breakdown complements the dimension-level aggregation reported in the main paper, revealing task-specific strengths and weaknesses for each video generation model.

\begin{table*}[t]
\centering
\small
\caption{Complete pass@5 results for all models across all 13 tasks in V-ReasonBench. Tasks are grouped by their reasoning dimensions.}
\label{tab:complete_results}
\begin{adjustbox}{width=\textwidth}
\begin{tabular}{llcccccc}
\toprule
\textbf{Dimension} & \textbf{Task} & \textbf{Seedance-1.0-Lite} & \textbf{Vidu-Q2} & \textbf{Kling-2.5-Turbo-Pro} & \textbf{Veo-3.1} & \textbf{Hailuo-02} & \textbf{Sora-2} \\
\midrule
\multirow{4}{*}{\makecell[l]{Structured\\Problem-Solving}} 
& Arithmetic Operation & 0.00 & 56.00 & 0.00 & 42.00 & 90.00 & 100.00 \\
& Code Execution & 0.00 & 0.00 & 2.22 & 11.11 & 31.11 & 53.33 \\
& Sudoku & 0.00 & 0.00 & 0.00 & 0.00  & 18.00 & 50.00 \\
& Tic-Tac-Toe & 3.33 & 3.33 & 26.67 & 63.33 & 46.67 & 90.00 \\
\midrule
\multirow{3}{*}{\makecell[l]{Spatial\\Cognition}} 
& Shape Fitting & 21.43 & 10.34 & 21.43 & 32.14 & 46.43 & 35.71 \\
& Visual Symmetry & 0.00 & 3.33 & 3.33 & 26.67 & 10.00 & 26.67 \\
& Color Connection & 0.00 & 0.00 & 20.00 & 10.00 & 60.00 & 60.00 \\
\midrule
\multirow{3}{*}{\makecell[l]{Pattern-based\\Inference}} 
& Sequence Completion & 0.00 & 10.00 & 10.00 & 5.00 & 15.00 & 0.00 \\
& Analogy Solving & 0.00 & 10.00 & 0.00 & 0.00 & 10.00 & 20.00 \\
& Rule Following & 0.00 & 24.00 & 0.00 & 20.00 & 40.00 & 48.00 \\
\midrule
\multirow{3}{*}{\makecell[l]{Physical\\Dynamics}} 
& Block Sliding & 20.00 & 20.00 & 10.00 & 20.00 & 10.00 & 20.00 \\
& Communicating Vessels & 0.00 & 10.00 & 0.00 & 10.00 & 20.00 & 0.00 \\
& Temperature-Induced Deformation & 80.00 & 90.00 & 60.00 & 90.00 & 100.00 & 60.00 \\
\bottomrule
\end{tabular}
\end{adjustbox}
\end{table*}

\section{Limitation of VLM-Based Evaluation}
\label{appendix:vlm}

\begin{figure}[!h]
    \centering
    \includegraphics[width=\linewidth]{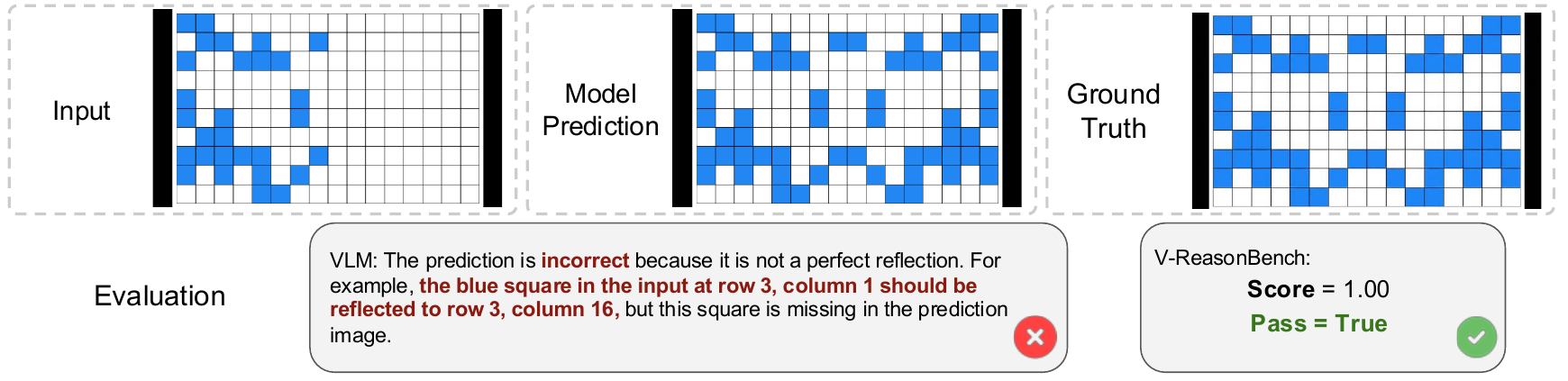}
    \caption{Example failure case from \textit{Visual Symmetry} task. Evaluation prompt for VLM: ``You are given three images: an input image, a ground-truth image, and a prediction image. The input image shows an original colored pattern. The correct transformation is: instantly reflect the entire pattern across its central vertical axis while preserving all existing colors and shapes exactly. No other modifications are allowed. Your task is to determine whether the prediction is correct. The prediction is correct only if two conditions are met: (1) The predicted pattern is a proper reflection of the input pattern across the central vertical axis, with no changes to colors, shapes, or structure other than mirroring. (2) The predicted pattern exactly matches the reflected pattern shown in the ground-truth image. This includes correct placement of every cell, color consistency, and no unintended alterations."}
    \label{fig:vlm_case1}
\end{figure}

\begin{figure}[!h]
    \centering
    \includegraphics[width=\linewidth]{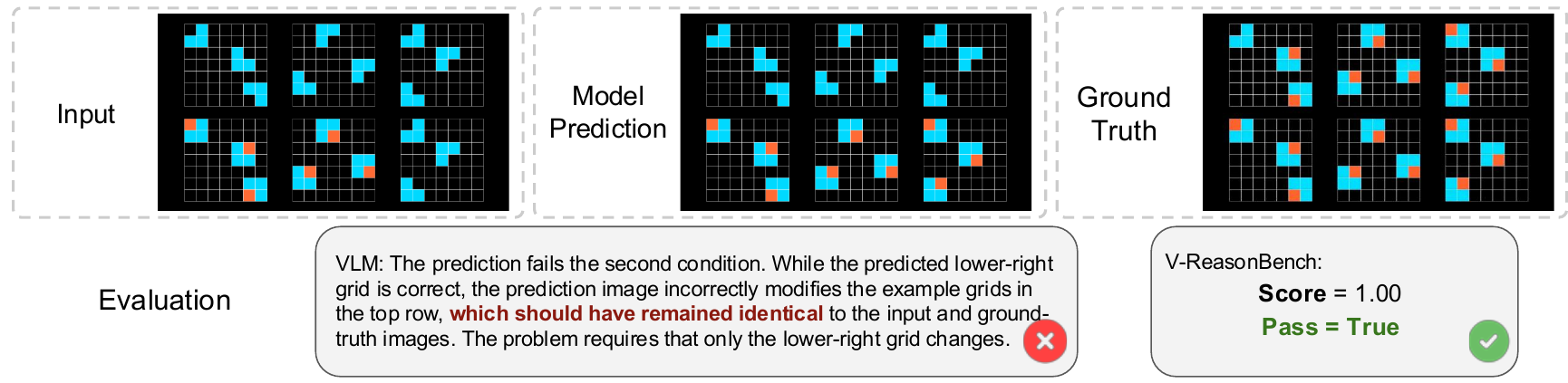}
    \caption{Example failure case from \textit{Rule Following} task. Evaluation prompt for VLM: ``You are given three images: an input image, a ground-truth image, and a prediction image. All images contain several example grids and a lower-right grid. Only the lower-right grid is supposed to change. Your task is to judge whether the prediction is correct.
The prediction is correct only if two conditions are met:
(1) The predicted lower-right grid follows the same transformation rule demonstrated by the example grids in the input image.
(2) The predicted lower-right grid visually and structurally matches the lower-right grid in the ground-truth image, including filled cells, empty cells, colors, and spatial arrangement. All other grids must remain identical across images."}
    \label{fig:vlm_case2}
\end{figure}

\begin{figure}[!h]
    \centering
    \includegraphics[width=\linewidth]{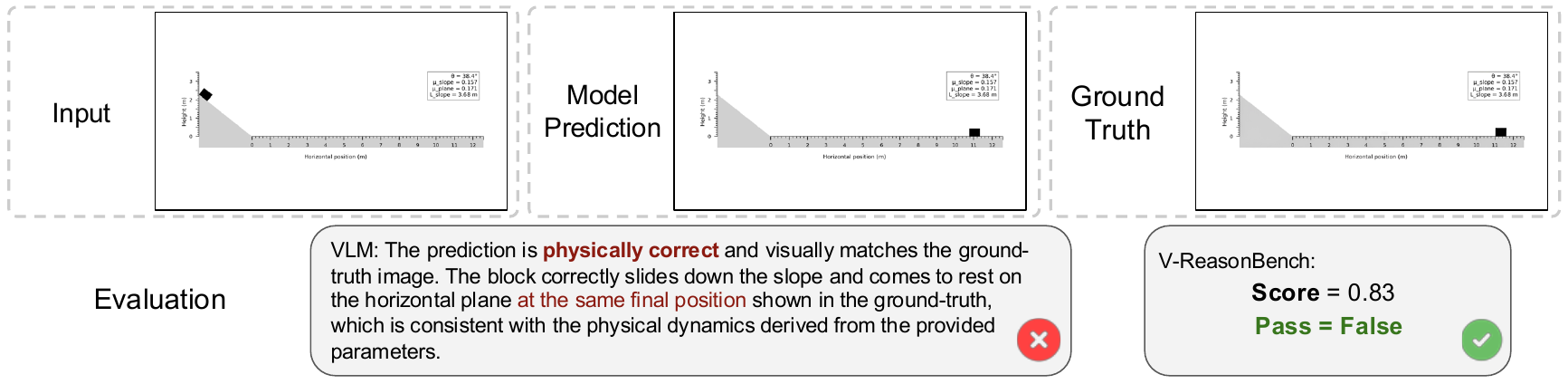}
    \caption{Example failure case from \textit{Block Sliding} task. Evaluation prompt for VLM: ``You are given three images: an input image, a ground-truth image, and a prediction image. The input image shows a rigid rectangular block resting at the top of an incline. The ground-truth image shows the final resting position of the block after sliding down the slope under ideal planar contact, given the annotated parameters: slope length $L_{slope}$, slope angle $\theta$, friction coefficients $\mu_{slope}$ and $\mu_{plane}$, block length and height, and gravitational acceleration $g = 9.81 m/s²$. Air resistance, elastic collisions, and secondary impacts at the slope foot are ignored. Your task is to determine whether the prediction is physically correct. The prediction is correct only if the following conditions are satisfied: (1) The predicted final resting position of the block is consistent with the physical dynamics implied by the slope geometry and friction parameters. The block must appear at the location where it should come to rest after sliding, without penetrating surfaces or hovering. (2) The predicted final frame visually matches the ground-truth image in block position, orientation, and contact location on either the slope or the flat plane. Colors, shapes, and scene layout must remain unchanged except for the block’s final location. (3) No extraneous motion or unintended scene modifications are introduced."}
    \label{fig:vlm_case3}
\end{figure}

Fig.~\ref{fig:vlm_case1}, \ref{fig:vlm_case2}, \ref{fig:vlm_case3} provide additional failure cases across multiple reasoning tasks. These examples highlight recurring issues in grid-structured and spatially dense scenes, where VLMs frequently misread fine-grained details such as cell boundaries, object adjacency, and subtle geometric relations.

\section{More Reasoning Patterns Demonstrations}
\label{appendix:reason}

To provide a clearer view of the structural deviation patterns discussed in the paper, we include additional example predictions of Seedance-1.0-Lite and Vidu-Q2 from the Tic-Tac-Toe task. These cases further illustrate how models differ in their handling of minimalist scenes in the video reasoning process that requires strict structural fidelity.

\begin{figure}
    \centering
    \includegraphics[width=\linewidth]{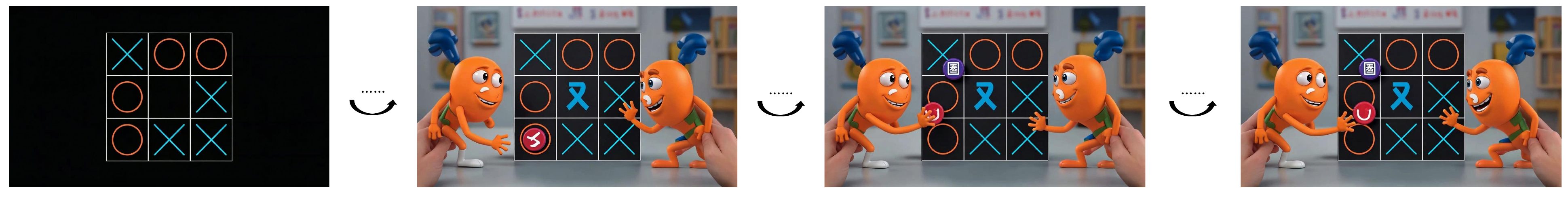}
    \caption{Given an initial board (left), Seedance-1.0-Lite generates a stylized video in which the board is embedded in a cartoon scene with characters and additional logos overlaid on several cells (right). Although the overall 3×3 grid and most X/O placements are preserved, the added icons and altered symbols change the board configuration, causing the prediction to fail under strict grid-based evaluation.}
    \label{fig:rp1}
\end{figure}

\begin{figure}
    \centering
    \includegraphics[width=\linewidth]{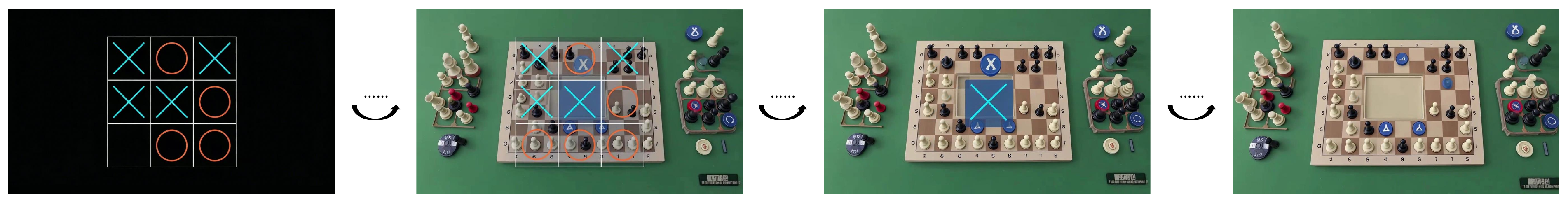}
    \caption{Starting from a simple target tic-tac-toe board (left), the model generates a chessboard scene in which the 3×3 grid is roughly aligned but many cells are filled with chess pieces and tokens instead of the required X and O marks. These changes alter the intended board configuration, so the prediction does not satisfy the 100\% cell-wise match required by the grid-based evaluation.}
    \label{fig:rp2}
\end{figure}

Taken together, these two examples in Fig. \ref{fig:rp1} and \ref{fig:rp2} show a consistent pattern in how Seedance-1.0-Lite handles simple board-based reasoning tasks. In both cases, the model starts from a clean tic-tac-toe target board and generates a richer, more realistic scene: in the first example, the board appears in a cartoon setting with characters and logo-like icons; in the second, the tic-tac-toe layout is mapped onto a chessboard surrounded by chess pieces and tokens. The model roughly keeps the 3×3 grid region and the idea of a board game, but it replaces several cells with new symbols or removes parts of the board entirely.

These changes are harmless from a storytelling point of view, but they break the strict structural requirements of the task. The final frames no longer contain the exact X/O pattern specified by the ground truth, so the predictions fail under our grid-based evaluation. These two cases illustrate how the model’s preference for visually rich, realistic scenes can conflict with reasoning tasks that depend on precise symbolic and spatial accuracy.

\begin{figure}
    \centering
    \includegraphics[width=\linewidth]{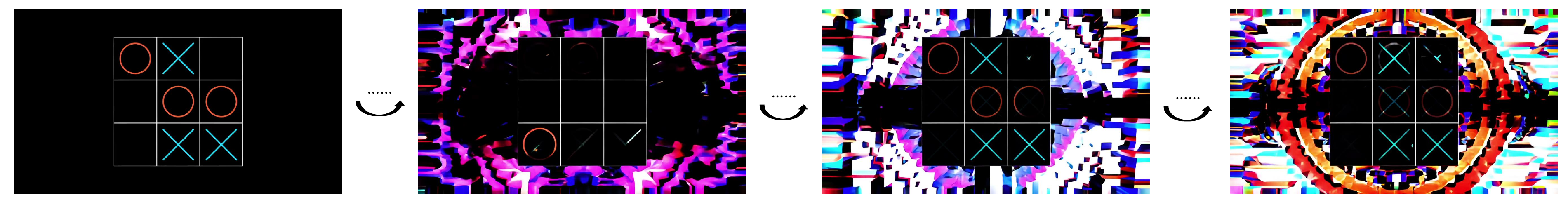}
    \caption{Starting from a simple initial board (left), the model generates a video in which the tic-tac-toe grid is placed on a colorful, highly textured background. Although the final frame roughly keeps the 3×3 grid and most X/O positions, some cells are partially overwritten by visual effects and altered marks.}
    \label{fig:rp3}
\end{figure}

\begin{figure}
    \centering
    \includegraphics[width=\linewidth]{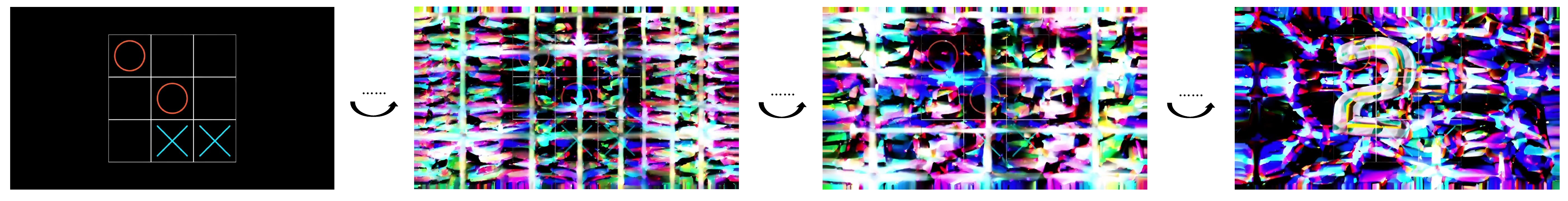}
    \caption{Starting from a defined initial board (left), the model generates frames in which strong abstract textures rapidly cover the scene. The tic-tac-toe grid and marks become barely visible and are eventually replaced by a large digit and background pattern.}
    \label{fig:rp4}
\end{figure}

Across these four tic-tac-toe examples, Seedance-1.0-Lite and Vidu-Q2 display the same tendency discussed in the main paper. When the input is a simple board on a plain background, the models transform it into a cartoon scene, a chessboard arrangement, or a heavily textured abstract image, instead of preserving the intended X/O pattern. Although they maintain a rough sense of a central board, the exact cell-level structure required for correct reasoning is lost.

This pattern is closely related to the models’ training data, which mostly consists of open-domain videos that emphasize rich visuals, motion, and diverse scenes, while containing very few clean, diagram-like examples. As a result, the models interpret minimal inputs as incomplete and replace them with visually complex content. However, video reasoning tasks require the opposite: models must treat simple scenes as complete and maintain precise spatial and symbolic structure. This indicates that training data and objectives for reasoning should place stronger weight on structure-preserving generation rather than visually complex outputs.

\section{More Hallucination in Video Reasoning}


\end{document}